%% file: main.tex
\documentclass[11pt]{article}

\usepackage[final]{acl}

\usepackage{times}
\usepackage{latexsym}

\usepackage{kotex}
\usepackage{amsmath}

\usepackage{enumitem}
\usepackage{booktabs}
\usepackage{tabularx}
\usepackage{booktabs}
\usepackage{tabularx}
\usepackage{array}
\usepackage{makecell}
\newcolumntype{Y}{>{\raggedright\arraybackslash}X}
\newcolumntype{P}[1]{>{\centering\arraybackslash}p{#1}}

\usepackage{xcolor}

\newcommand{\gold}[1]{\textbf{\textcolor{blue}{#1}}}

\usepackage[T1]{fontenc}

\usepackage[utf8]{inputenc}

\usepackage{microtype}
\usepackage{inconsolata}

\usepackage{graphicx}

\title{KoNeoBench: A Curated Evaluation Dataset for \\ LLM Understanding of Korean Neologisms}
\author{
  \textbf{Soha Lee}\textsuperscript{1}\thanks{These authors contributed equally to this work.},
  \textbf{Soojin Lee}\textsuperscript{2}\footnotemark[1],
  \textbf{Heesung Yang}\textsuperscript{1},
  \textbf{Hyunju Song}\textsuperscript{3},\\
  \textbf{Hyunji Lee}\textsuperscript{1},
  \textbf{Jinsan An}\textsuperscript{4},
  \textbf{Jeongwan Shin}\textsuperscript{5},
  \textbf{Jin Hyun Park}\textsuperscript{6},
  \textbf{Jun Lee}\textsuperscript{7},\\
  \textbf{Hyeyoung Park}\textsuperscript{1}\thanks{Corresponding author: Kilim Nam (\texttt{nki@yonsei.ac.kr}), Hyeyoung Park (\texttt{hypark@knu.ac.kr})},
  \textbf{Kilim Nam}\textsuperscript{7}\footnotemark[2]
  \\
  \textsuperscript{1}School of Computer Science and Engineering, Kyungpook National University \\
  \textsuperscript{2}International Exchange Department, Kyungpook National University \\
  \textsuperscript{3}Dept. of Korean Language and Literature Education, Kyungpook National University \\
  \textsuperscript{4}Dept. of Korean Language and Literature, Kyungpook National University \\
  \textsuperscript{5}Daegu Gyeongbuk Institute of Science and Technology (DGIST) \\
  \textsuperscript{6}Dept. of Computer Science and Engineering, Texas A\&M University, \\
  \textsuperscript{7}Dept. of Korean Language and Literature, Yonsei University
  \\
}

\begin{document}
\maketitle
\begin{abstract}
Large language models (LLMs) are typically evaluated on static benchmarks, even though natural language constantly evolves through newly emerging words and meanings. Existing Korean benchmarks are centered on established vocabulary and therefore provide limited coverage of such recent lexical change, and their English-oriented design makes it difficult to assess the typological properties of Korean, in which content words combine productively with functional morphemes. In this paper, we introduce \textbf{KoNeoBench}, a benchmark for evaluating LLMs' understanding of Korean neologisms. KoNeoBench is built on 1,785 Korean neologisms attested in online news since 2020 and curated through expert lexicographic review. Each entry provides usage examples, word-formation analyses, and dictionary-style definitions. Based on this resource, we define four tasks and report results on recent models, together with a human baseline. Our experiments show that current LLMs exhibit clear limitations in recovering source components, distinguishing semantic categories, and generating accurate definitions. 
These results reveal specific aspects of recent Korean lexical change that remain challenging for current LLMs.
KoNeoBench is available at \href{https://github.com/bcmilab/ko-neobench/}
\texttt{https://github.com/bcmilab/ko-neobench/}.
\end{abstract}

\section{Introduction}
Large language models (LLMs) achieve strong performance on many language tasks, but their evaluation still relies largely on static benchmarks built from fixed linguistic data and knowledge, 
\begin{figure}[h]
  \centering
  \includegraphics[width=1.0\columnwidth]{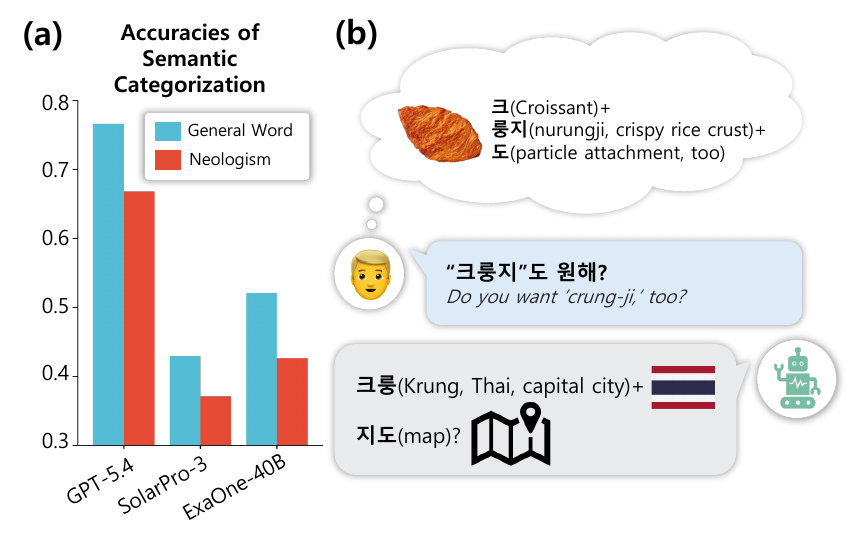}
  \caption{Challenges in understanding Korean neologisms. (a) Semantic classification performances of LLMs neologisms vs. general words. (Details are given in Appendix \ref{app:C.2}.) (b) A neologism formed by abbreviation and blending.}
  \label{fig:fig1}
  \vspace{-0.5cm}
\end{figure}
which may not fully capture aspects of recent language change.
Prior work has highlighted this limitation in terms of temporal generalization and temporal misalignment, showing performance degradation on future text, time-sensitive facts, and updated world knowledge \citep{lazaridou2021mind, luu2022time, dhingra2022time, vu2024freshllms, zhu2025your}.
Yet 
recent language change is not limited to factual updates;
it also emerges at the lexical and semantic levels through new words and shifts in meaning. As early indicators of such change, neologisms reflect not only changes in linguistic form and meaning but also the social and cultural context of the language community \citep{nam2025neologism}. 
Accordingly, evaluating an LLM’s understanding of neologisms can provide insight into how well it captures recent lexical and sociocultural developments in language use.

Recent studies have begun to evaluate LLMs’ adaptation to language change through neologisms and informal expressions. NeoBench \citep{zheng2024neobench} introduces benchmark data and tasks for neologism processing, while \citet{mei2024slang} and \citet{sun2024toward} examine LLMs’ understanding, detection, and source identification of internet slang, memes, and informal expressions. 
However, these studies focus primarily on English, and benchmarks for Korean neologisms remain limited.
Within Korean, our preliminary results in Figure~\ref{fig:fig1}(a) also reveal a substantial performance gap between established words and neologisms across different LLMs, motivating a closer examination of their understanding of Korean neologisms.

Beyond this performance gap, Korean neologisms exhibit linguistic characteristics that warrant explicit consideration in benchmark design.
Korean is an agglutinative language in which particles and inflectional endings attach to lexical stems. Figure~\ref{fig:fig1}(b) illustrates how a neologism formed through abbreviation and/or blending can combine with a grammatical morpheme, requiring models to identify both the neologism boundary and the function of the attached morpheme. Korean also contains many hybrid neologisms formed from multiple lexical strata, including native Korean, Sino-Korean, and loanwords. These properties make Korean neologisms particularly difficult to interpret and analyze, and they should therefore be taken into account when evaluating model understanding.

To address the limited coverage of Korean neologisms in existing benchmarks, we introduce \textbf{KoNeoBench}, a benchmark for multifaceted evaluation of LLMs’ understanding of Korean neologisms. It is built from a list of Korean neologisms collected and annotated through expert lexicographic review. It comprises four tasks: (1) Cloze task for contextual understanding of neologism usage, (2) source word identification aligned with native speakers’ interpretation, (3) semantic and specialized domain categorization, and (4) dictionary-style definition generation. Using KoNeoBench, we evaluate multiple LLMs and analyze their strengths and limitations in processing Korean neologisms from various perspectives.
The main contributions of this paper are as follows.
\begin{itemize}
\item We present \textbf{KoNeoBench}, a Korean neologism dataset compiled and annotated by lexicography experts.
\item \textbf{KoNeoBench} is the first benchmark to provide carefully designed test sets for multi-faceted evaluation of LLMs’ understanding of Korean neologisms.
\item Using these Korean-specific test sets, we evaluate a range of LLMs and analyze the major challenges and error patterns involved in processing Korean neologisms.
\end{itemize}

\section{Related Works}
\label{app:related_works}

\paragraph{Temporal Drift and Robustness in LLMs}
Research on the robustness of LLMs under temporal change has primarily examined performance degradation on post-training data \citep{lazaridou2021mind, luu2022time} and time-sensitive downstream tasks such as question answering \citep{chen2021dataset, liska2022streamingqa, dhingra2022time,
agarwal2022temporal}. Other studies have shown that emerging entities and shifts in word meaning or frequency can likewise affect model performance \citep{rijhwani2020temporally, onoe2022entity, pramanick2022challenges}. Together, these findings motivate the study of neologisms as a lexical-level manifestation of temporal language change.

\paragraph{Neologism Research in LLMs}
Research on neologisms has traditionally focused on collecting and classifying new words, analyzing their formation and emergence, and
tracing their diffusion in online communities \citep{pinter2020nytwit, ryskina2020where, zhu2021structure}. More recently, studies have begun to evaluate how well LLMs understand rapidly changing expressions such as neologisms, slang, and memes \citep{zheng2024neobench, sun2024toward, mei2024slang}. Unlike traditional work on out-of-vocabulary (OOV) items, this line of research emphasizes novel forms and meanings, word-formation patterns, and community-based usage \citep{garneau2018predicting, nayak2020domain, araabi2022how}. Earlier computational studies, for example, investigated the recovery of source words in English lexical blends and the incorporation of linguistic knowledge about neologism formation \citep{cook2010automatically, cook2011exploiting}, while recent work
has expanded to lexical proficiency, scientific neologism translation, slang sense alternation, and culturally or regionally specific slang \citep{ciaccio2025evaluating, lerner2025towards, aloraini2026slangtrack, wu2025how, wuraola2026slanggraphrag}.

Among these studies, NEO-BENCH \citep{zheng2024neobench} is the most closely related to our work. It evaluates recent English neologisms through machine translation, cloze question answering, definition generation, and perplexity, with an emphasis on model robustness to lexical temporal drift. KoNeoBench complements this perspective by focusing on Korean lexical change and evaluating multiple aspects of neologism knowledge, including neologism  recognition, source-component recovery, semantic and specialized-domain categorization, and definition generation. In particular, its expert-curated lexical annotations allow us to evaluate not only whether a model recognizes or interprets a recent expression in context, but also whether it captures the word-formation and semantic properties that characterize Korean neologisms.

\paragraph{Korean NLP Benchmarks and Neologism}
Korean NLP has also seen the development of various evaluation resources. Benchmarks such as KLUE, KoBEST, HAE-RAE Bench, and Ko-H5 provide important foundations for evaluating understanding, reasoning, knowledge, cultural context, and LLM performance \citep{park2021klue, jang2022kobest, son2024haerae, park2024open}. However, they mainly target general language understanding or relatively static linguistic knowledge. In parallel, linguistic research on Korean neologisms has examined not only data collection but also word-formation patterns, usage contexts, and lexicalization processes \citep{moon1999types, lee2014extraction, nam2015analysis, lee2006contextual, lee2024corpus}. Building on these two lines of research, KoNeoBench provides a linguistically grounded benchmark to evaluate how well LLMs understand the recent Korean lexical change across morphological, semantic, and contextual dimensions.

\section{KoNeoBench: A Benchmark Dataset of Korean Neologisms}

\subsection{Data Collection}
\label{sec:data_collection}
\begin{figure}[h]
    \centering
    \includegraphics[width=0.85\columnwidth]{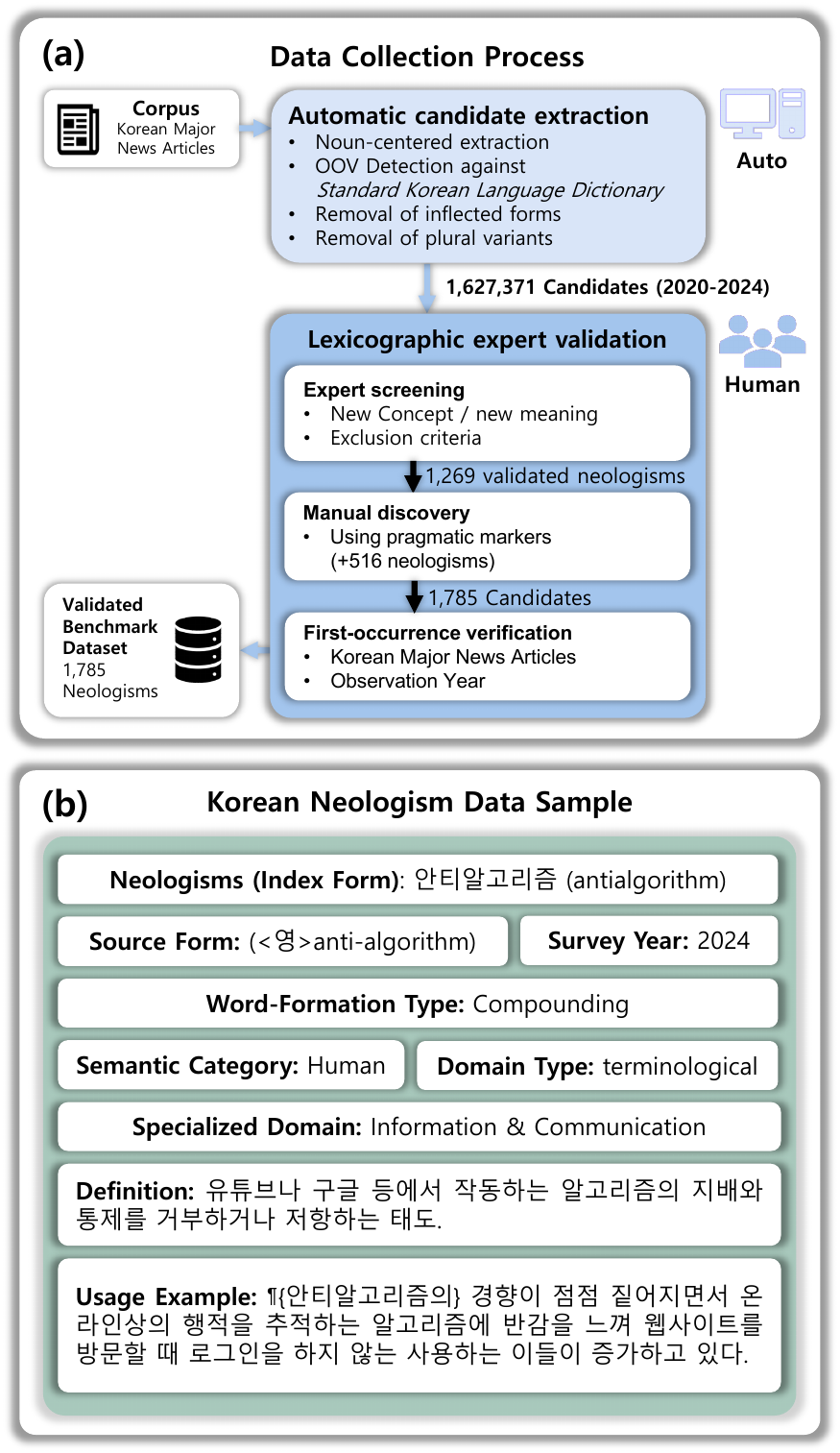}
    \vspace{-0.2cm}
    \caption{KoNeoBench dataset: (a) data collection process and (b) an example of a neologism and its attributes.}
    \label{fig:benchmark_construction}
    \vspace{-0.3cm}
\end{figure}

Korean neologism data have been compiled annually under the supervision of the National Institute of Korean Language since 1994. We use 1,785 neologisms listed in annual publications produced since 2020 under the same methodology. The source corpus consists of articles from major general and business newspapers in Korea. Candidate neologisms were first extracted automatically as out-of-vocabulary items, and were then screened and supplemented by lexicography experts. The final list for each year was determined by verifying first occurrence dates in \textit{Naver News}, which contains article data from around 130 media outlets. Figure \ref{fig:benchmark_construction}(a) illustrates the overall collection pipeline, and Appendix \ref{app:A.1} provides additional details.

\begin{figure*}[h]
    \centering
    \includegraphics[width=1.0\textwidth]{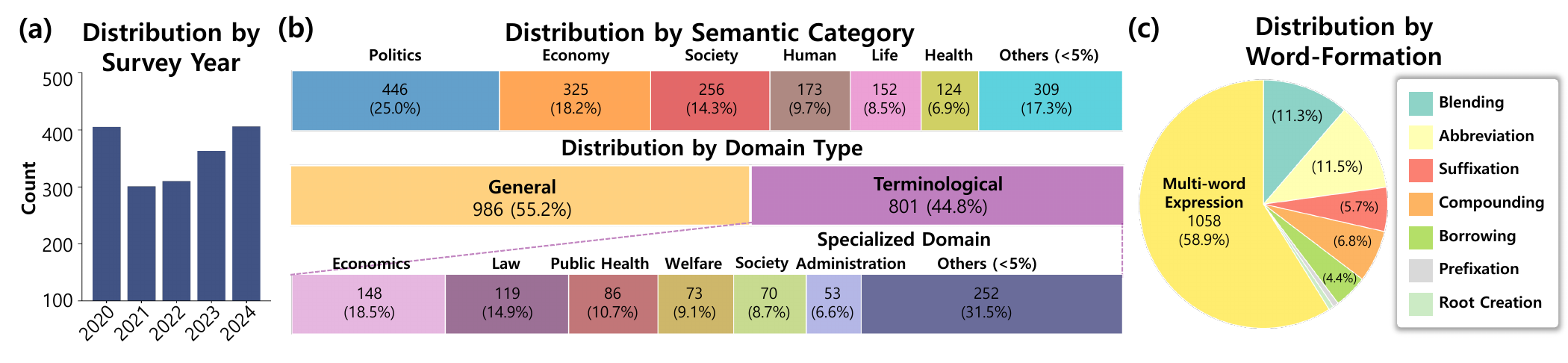}
    \vspace{-0.5cm}
    \caption{Data Statistics in KoNeoBench. (a) Distribution by survey year. (b) Distribution by semantic and specialized domain categories. (c) Distribution by type of word-formation.}
    \label{fig:benchmark_statistics}
    \vspace{-0.3cm}
\end{figure*}
\vspace{-0.2cm}

In the automatic candidate collection stage, candidate words are identified mainly from nouns by comparing corpus items with the headword list of the \textit{Standard Korean Language Dictionary}, the largest normative dictionary of Korean, and then removing inflected and plural forms. Because Korean is an agglutinative language, this process inevitably produces noise and may miss genuine neologisms, making expert review essential. 

In the validation stage, candidates are first filtered using rule-based criteria and then reviewed by experts.
A candidate is accepted as a neologism if it (1) is created to refer to a newly emerged object or phenomenon, (2) expresses a new nuance despite the existence of a conventional expression, or (3) functions as a phrasal unit with a unified lexical meaning. Proper nouns, group-specific expressions, nonce words, and foreign words not established in the Korean lexical system are excluded. In practice, the automatically extracted out-of-vocabulary list contained 1,627,371 candidates over five years, but most were mechanically identified errors such as incomplete fragments, failed morphological analyses, or misspellings. After automatic filtering and expert review, only 1,269 items were retained as neologisms, illustrating the substantial reduction required to obtain the final set.

In addition, to find neologisms that could not be captured by automatic extraction, we conducted a manual addition stage by experts using pragmatic markers in the corpus, through which 516 additional neologisms were obtained. Pragmatic markers are contextual cues in text that signal the presence of a neologism, as in expressions such as the Korean expression `$x$-라는 신어' (`a neologism called $x$'), which are regarded as important indicators for assessing both the novelty and the degree of establishment of a neologism \citep{klosa2020considerations, lee2024corpus}.

\subsection{Data Descriptions and Statistics}
As shown in Figure \ref{fig:benchmark_construction}, KoNeoBench includes not only a list of neologisms but also their associated attributes: dictionary-style definitions and usage examples, survey year, the source form, the word-formation type, the domain type (general vs. terminological), the semantic category, and the specialized domain for terminological neologisms. Detailed descriptions of these attributes are provided in Appendix \ref{app:A.2}.

Figure \ref{fig:benchmark_statistics} presents the distributional characteristics of the dataset from several perspectives. Across the years, roughly 400 neologisms were consistently collected each year from 2020 to 2024. In terms of word-formation patterns, multi-word (or phrasal) neologisms formed by combining two or more words, account for the largest share at 58.9\%, followed by neologisms created through abbreviation and blending. This pattern is closely related to the spread of English through the web. The high proportion of abbreviated and blended forms, such as `호캉스 (ho-kangseu, `hotel + vacance,' a staycation at a hotel), also reflects a characteristic of Korean word-formation that allows relatively free truncation and recombination at the syllable level, particularly in loanwords from English.

The distribution in Figure \ref{fig:benchmark_statistics}(b) shows that the top six semantic categories (politics, economy, society, human, life, and health) account for 82.7\% of the dataset, suggesting that these areas are socially salient domains in which new vocabulary is actively created and circulated within the community. A similar skew is observed in the distribution of the 801 terminological neologisms, with particularly high proportions in health and welfare, reflecting the exceptional social conditions of the COVID-19 pandemic. 
In this way, neologisms reflect the sociocultural context of the period in which they emerge. Evaluating LLMs on neologisms therefore measures not only lexical knowledge but also whether models recognize lexical expressions tied to recent social and cultural phenomena. Direct evaluation of broader sociocultural reasoning is left to future work.

\subsection{Evaluation Tasks}
\begin{figure*}[h]
    \centering
    \includegraphics[width=1.0\textwidth]{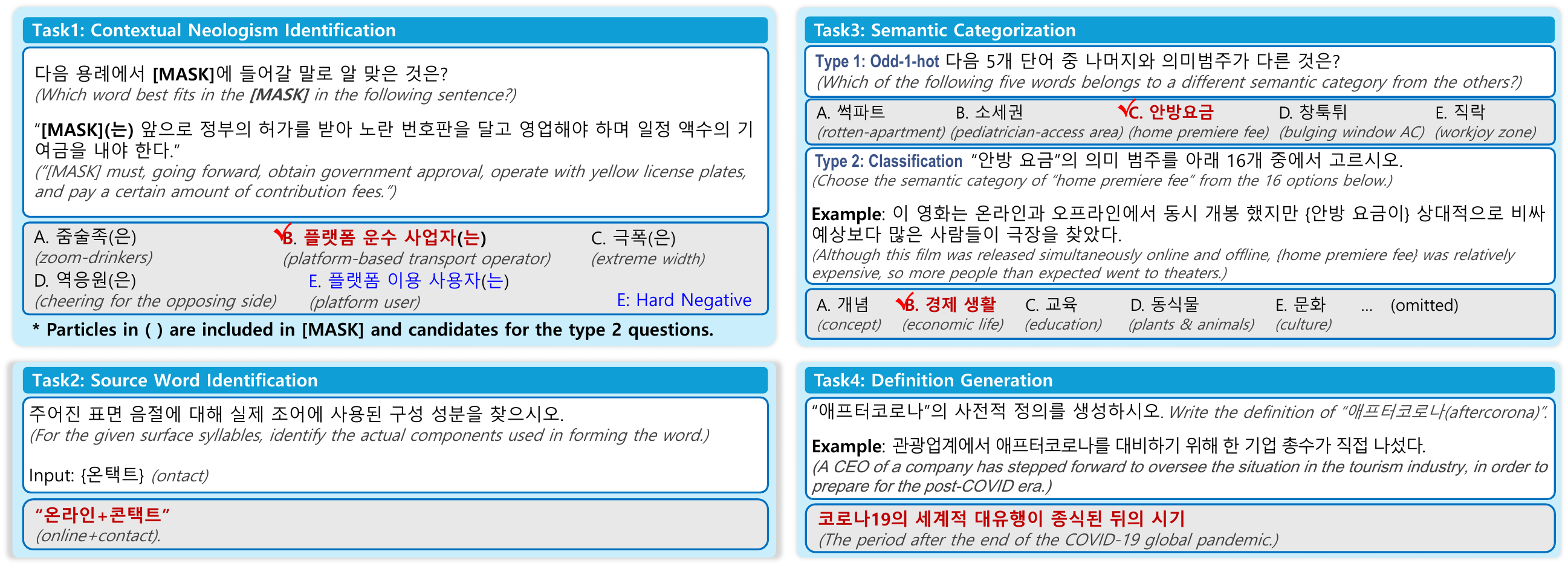}
    \vspace{-0.5cm}
    \caption{Examples of the four evaluation tasks in KoNeoBench.}
    \vspace{-0.5cm}
    \label{fig:benchmark_tasks}
    
\end{figure*}

KoNeoBench provides not only a list of neologisms but also associated attributes, which make it possible to design a variety of evaluation tasks. In this paper, we designed four tasks for evaluating contextual understanding, source word identification, categorical understanding, and definition generation in neologism processing, and provide test sets consisting of questions, gold answers, and evaluation criteria. Figure \ref{fig:benchmark_tasks} presents example instances for each task, and Appendix \ref{app:B} provides the details.

\subsubsection{Task 1: Contextual Identification}
Task 1 is a multiple-choice cloze task in which the masked neologism must be identified from its original usage context. Distractors include contextually plausible but incorrect hard negatives constructed based on lexical semantic relations, such as conceptual or morphological similarity \cite{leech1981semantics}, as well as randomly selected words that serve as a baseline. Additionally, because Korean case particles can provide semantic and syntactic cues, we construct two variants of the test set: Type 1, where the mask and answer candidates contain only the word, and Type 2, where the particle is included together with the word (See Figure \ref{fig:benchmark_tasks}).

\subsubsection{Task 2: Source Word Identification}
Task 2 is motivated by Korean lexicographic practice, which marks component boundaries in complex words, and by speakers' tendency to infer unfamiliar neologisms through their constituents. Even when encountering an unfamiliar word not listed in a dictionary, native speakers can segment its source words and infer its meaning by identifying the underlying word-formation process, such as compounding, abbreviation, or blending. This task evaluates whether LLMs can perform a similar kind of linguistic inference. 

The test set covers 396 neologisms whose formation types are blends or abbreviations, and includes gold annotations of their constituents and corresponding source words. These forms combine truncated or omitted source words, making the original forms difficult to identify from the surface form alone. For example, in `danmato → dalda(sweet) + tomato', the source words cannot be reconstructed simply by segmenting the surface syllables. To solve this task, an LLM must use the partial syllabic form of a neologism as a cue and identify semantically and phonologically plausible source words in the Korean lexicon. 
The task therefore evaluates source-component recovery and sensitivity to recurring formation patterns, beyond whether the model has learned the surface form and meaning of neologisms.
We use prompts that explicitly instruct the model to identify the source-component recovery; examples are provided in Appendix \ref{app:B.2}.

\subsubsection{Task 3: Neologism Categorization}
Task 3 evaluates whether a model can understand the categorical meaning of Korean neologisms. It consists of two subtasks: semantic categorization, which assigns each neologism to one of 16 general semantic categories, and specialized domain categorization, which assigns 801 terminological neologisms to one of 43 specialized domains.

Semantic categorization tests whether a model can classify neologisms into semantic categories as humans do, reflecting the fact that neologisms typically emerge within particular semantic domains. The specialized domain categorization task asks whether the model can identify the specialized domain of a terminological neologism, thereby measuring its sensitivity to register and domain specificity.
Since terminological neologisms are often tied to particular knowledge communities and may be opaque outside them, this task provides a useful indicator of whether a model can capture not only its basic meaning but also its domain of use.

For each category system, we design two question formats; Type 1 is an odd-one-out task in which the model selects the one item that belongs to a different category from five candidate neologisms; Type 2 is a classification task in which the model directly predicts the category of a given neologism (See Figure \ref{fig:benchmark_tasks}). 
By evaluating both tasks, it is possible to distinguish whether a model can merely group words into a coherent lexical class or can further link that class to an explicit category label. A neologism may be grouped with other words from the same class based only on formal similarity or co-occurring context, but assigning the correct category name requires a different level of understanding. For Type 2, we evaluate two input settings: one with the neologism alone, and the other with both the neologism and its usage example.

\subsubsection{Task 4: Definition Generation}
Task 4 is a generative task that evaluates whether a model can produce a dictionary-style definition of a neologism. It goes beyond identifying a neologism’s form or category and tests whether the model can explain its core meaning in natural language. We consider two input settings: a context-free setting with only the neologism, and a context-based setting with both the neologism and a usage example.

Definition writing is one of the most difficult tasks in lexicography, as it requires both semantic analysis and concise expression in a restricted metalanguage\cite{hanks2015definition}. This makes the task linguistically meaningful, as it evaluates whether a model can both infer the meaning of a neologism and express it accurately in dictionary-style language. The prompt template is provided in Appendix \ref{app:B.2}.

\section{Evaluation and Analysis}
\subsection{Evaluation models}
Using KoNeoBench, we evaluate recent LLMs and analyze the results. Our goal is not to compare the superiority of particular model families, but to examine the overall performance trends and limitations of LLMs with different training backgrounds, scales, release dates, and degrees of suitability for Korean language processing in handling Korean neologisms. To this end, we evaluate a total of nine models, including commercial API-based models, open-weight models, and Korean-specialized models. Details of the evaluated models are summarized in Table \ref{tab:evaluation_models} in the Appendix.

\subsection{Evaluation criteria}
For Task 1 and Task 3, we use the standard evaluation metric of classification accuracy, since both are multiple-choice tasks. For Task 2 and Task 4, which require free-form responses, we design task-specific evaluation criteria.

\paragraph{Coverage \& Precision for Task 2.}
To evaluate source word identification in Task 2, we use coverage, precision, and their harmonic mean (F1) as metrics, so as to account for cases where the model identifies only part of the gold source words or generates unnecessary additional ones. Coverage measures the proportion of gold source words correctly identified by the model, while precision measures the proportion of generated source words that are actually included in the gold answer. Details are provided in Appendix \ref{app:B.3}.

\paragraph{LLM-as-a-Judge for Task 4.}
We evaluate LLMs' generated definitions by comparing them with the gold definitions in KoNeoBench. Because the same meaning can be expressed in multiple ways, simple lexical matching is not appropriate. We therefore adopt an LLM-as-a-judge approach, using Gemini, a model from a different family than the evaluated LLMs, to assess semantic equivalence.

LLM-based evaluation of generated outputs has become common and has also been adopted in prior benchmark studies \cite{zheng2023judging, zheng2024neobench}. To address concerns about reliability, we propose evaluation criteria carefully designed by lexicography experts. The criteria cover three dimensions for a total of 10 points: semantic adequacy, fluency, and factuality. The full criteria and judge prompt are given in Appendix \ref{app:B.4} and \ref{app:B.2}. We also conduct human validation on a subset of samples to verify the reliability of the automatic evaluation; details are provided in the results section.

\subsection{Evaluation Results and Analysis}
\begin{table*}[t]

\centering
\small
\setlength{\tabcolsep}{3.2pt}
\renewcommand{\arraystretch}{1.15}
\caption{Evaluation results on four tasks in KoNeoBench. (Ave. pts. denotes the average LLM-as-a-judge score points. For each task, the best performance is shown in red and the worst performance in blue. )}
\label{tab:overall_results}
\resizebox{\textwidth}{!}{%
\begin{tabular}{lccccccclcclcc}
\toprule
Model
& \multicolumn{2}{c}{Task 1 (Acc. \%)}
& \multicolumn{3}{c}{Task 2} 
& \multicolumn{3}{c}{Task 3 Semantic (Acc. \%)}
& \multicolumn{3}{c}{Task 3 Spec. Domain (Acc. \%)}
& \multicolumn{2}{c}{Task 4 (Ave. pts)} \\
\cmidrule(lr){2-3}
\cmidrule(lr){4-6}
\cmidrule(lr){7-9}
\cmidrule(lr){10-12}
\cmidrule(lr){13-14}

& type 1
& type 2
& cov.
& prec.
& F1
& type 1
& \multicolumn{2}{c}{type 2}
& type 1
& \multicolumn{2}{c}{type 2}
& w/o ex.
& w/ ex. \\
\cmidrule(lr){8-9}
\cmidrule(lr){11-12}

& w/o part. & w/ part. & & & &
& w/o ex. & w/ ex.(diff) 
& & w/o ex. & w/ ex.(diff) 
& & \\
\midrule
GPT-4.1
& 54.62 & 52.21
& 43.71 & 44.13 & 43.78
& 63.00 & 47.00 & 57.54(10.54$\uparrow$) 
& 80.00 & 57.88 & 63.38(5.50$\uparrow$) 
& $7.15$ & $8.47$ \\

GPT-5.4
& \textcolor{red}{\textbf{66.16}} & \textcolor{red}{\textbf{62.97}}
& \textcolor{red}{\textbf{49.77}} & \textcolor{red}{\textbf{50.33}} & \textcolor{red}{\textbf{49.96}}
& \textcolor{red}{\textbf{66.80}} & \textcolor{red}{\textbf{53.28}} & \textcolor{red}{\textbf{62.52}}(\textcolor{blue}{\textbf{9.24}}$\uparrow$) 
& \textcolor{red}{\textbf{84.12}} & \textcolor{red}{\textbf{64.38}} & \textcolor{red}{\textbf{67.50}}(\textcolor{blue}{\textbf{3.12}}$\uparrow$) 
& \textcolor{red}{$\bf 7.49$} & \textcolor{red}{$\bf8.69$} \\

Solar Pro3-12B
& 44.34 & 41.82
& 32.48 & 33.86 & 32.90
& 37.14 & 36.13 & 49.52(13.39$\uparrow$) 
& 52.12 & 49.38 & 57.25(7.87$\uparrow$) 
& $6.68$ & $7.79$ \\

Solar-10.7B
& 37.98 & 36.69
& \textcolor{blue}{\textbf{9.89}} &\textcolor{blue}{\textbf{9.87}} & \textcolor{blue}{\textbf{9.76}}
& 32.77 & \textcolor{blue}{\textbf{25.10}} & \textcolor{blue}{\textbf{37.98}}(12.88$\uparrow$) 
& 49.50 & 23.25 & 33.12(9.87$\uparrow$) 
& \textcolor{blue}{$\bf 3.5$} & \textcolor{blue}{$\bf 4.41$} \\

EXAONE-3.5-7.8B
& 47.09 & 45.40
& 23.48 & 24.93 & 24.00
& 34.12 & 33.45 & 48.12(14.67$\uparrow$) 
& 40.12 & \textcolor{blue}{\textbf{17.00}} & \textcolor{blue}{\textbf{22.38}}(5.38$\uparrow$) 
& $6.17$ & $7.53$ \\

EXAONE-4.0-32B
& 52.86 & 50.84
& 23.80 & 24.69 & 24.08
& 42.69 & 37.14 & 53.05(15.91$\uparrow$) 
& 54.62 & 50.38 & 63.00(12.62$\uparrow$) 
& $6.07$ & $7.53$ \\

Qwen2.5-7B
& 40.47 & 38.00
& 18.03 & 18.79 & 18.19
& 36.53 & 34.62 & 48.80(14.18$\uparrow$) 
& 48.00 & 42.50 & 53.50(11.00$\uparrow$) 
& $5.56$ & $6.81$ \\

Qwen3.5-9B
& 47.40 & 46.64
& 23.42 & 24.04 & 23.62
& 39.94 & 37.76 & 51.09(13.33$\uparrow$) 
& 57.13 & 48.25 & 57.88(9.63$\uparrow$) 
& $5.58$ & $7.05$ \\

LLaMA-3.1-8B
& \textcolor{blue}{\textbf{34.47}} & \textcolor{blue}{\textbf{30.72}}
& 15.51 & 15.80 & 15.50
& \textcolor{blue}{\textbf{28.90}} & 32.94 & 47.68(14.74$\uparrow$) 
& \textcolor{blue}{\textbf{32.75}} & 33.25 & 39.62(6.37$\uparrow$) 
& $5.19$ & $6.31$ \\
\bottomrule

Mean
& 47.27 & 45.03
& 26.68 & 27.38 & 26.87
& 42.43 & 37.00 & 50.70(13.70$\uparrow$) 
& 55.37 & 42.92 & 50.85(7.93$\uparrow$) 
& $5.93$ & $7.18$ \\


Max$-$Min
& 31.69 & 32.25
& 39.88 & 40.46 & 40.20
& 37.90 & 28.00 & 24.54 
& 51.37 & 47.38 & 45.12 
& $3.99$ & $4.28$ \\
\bottomrule
\end{tabular}%
}
\end{table*}
Table \ref{tab:overall_results} summarizes the results of all models across all tasks. The GPT series performs best overall, while earlier open models such as Solar-10.7B and LLaMA-3.1 perform worst, likely reflecting the time-sensitive nature of neologisms. In most tasks, average scores remain around 30–50 out of 100, indicating that LLMs’ understanding of Korean neologisms is still limited.
Since most evaluated models were released between 2024 and 2026 (See Table \ref{tab:evaluation_models}), after the emergence of the large portion of neologisms in KoNeoBench, this low performance cannot be attributed solely to knowledge cutoff and may also reflect the inherently low-frequency nature of neologisms.
Performance gaps are especially large in Task 4, suggesting that definition generation is particularly discriminative. Large variation is also observed in the specialized domain categorization task (Task 3), possibly due to differences in training data diversity. Figure \ref{fig:Task_wise_scatter_plot} demonstrates further information for each task.
We also compared model performance with human performance on subsets of Tasks 1--3. Linguistics majors performed better than GPT-5.4 on Tasks 1 and 2, while GPT-5.4 performed better than the overall human group on the specialized-domain setting of Task 3. Full results and statistical tests are reported in Appendix~\ref{app:C.3.human_eval}.

\subsubsection{Results on Task 1} 
As shown in Table~\ref{tab:overall_results}, the best performance is achieved by GPT-5.4, a recently released API model. GPT-4.1, released about a year earlier, and EXAONE-4.0, a Korean-specialized model released around the same time, show similar performance. By contrast, Solar Pro3, despite being a recently released Korean-specialized model, performs below expectations, even falling behind EXAONE-3.5, a smaller Korean-specialized model released earlier.

\begin{figure}[h]
    \centering
    \includegraphics[width=\columnwidth]{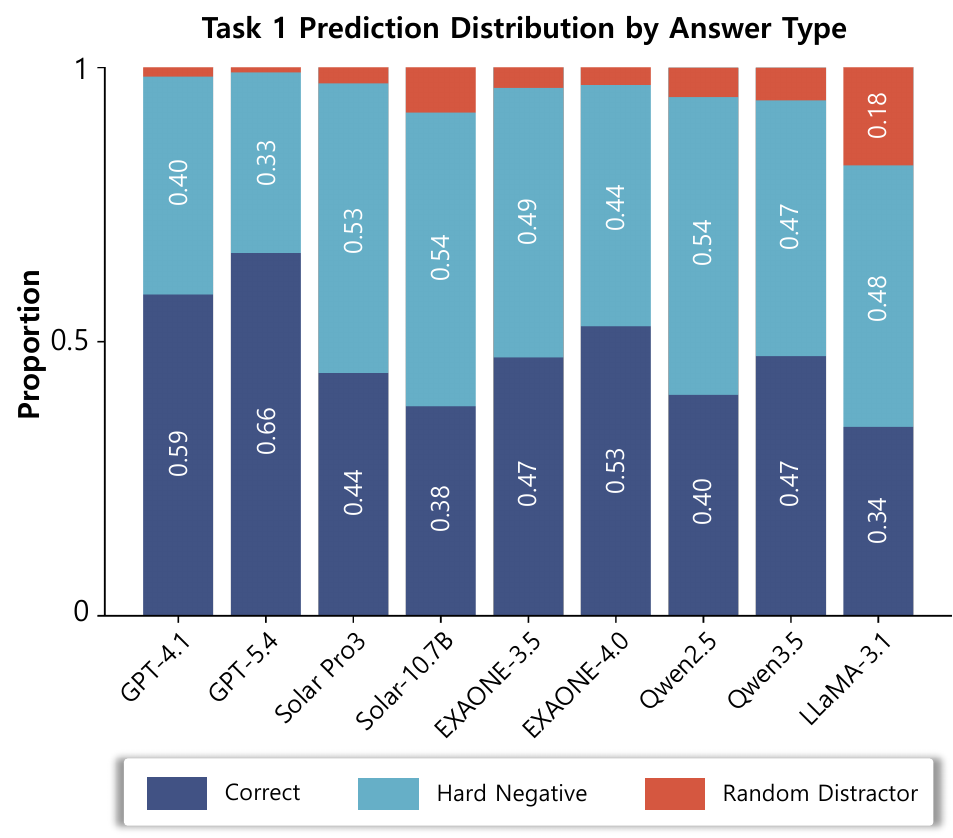}
    \caption{Distribution of LLM responses to type 1 questions in Task 1.}
    \label{fig:task1_type1_dist}
    \vspace{-0.3cm}
\end{figure}

Even the best-performing model reaches only the 60\% range in accuracy. To better understand this limitation, Figure~\ref{fig:task1_type1_dist} breaks down the types of answers selected by the models. Most incorrect predictions are hard negatives, showing that this task measures not only general contextual understanding but also fine-grained discrimination among neologisms. In other words, models can usually identify the broad contextual area in which a neologism should appear, but often fail to distinguish the correct neologism from other plausible candidates within that area. The poor performance of the Korean-specialized model Solar Pro3 is also largely due to these hard-negative errors (53\%), suggesting that learning general Korean vocabulary alone is not sufficient for fine-grained discrimination among neologisms. 

We also compare the results by whether the masked item includes a case particle. For all models, Type 2, which includes the particle, performs worse than Type 1 (average −2.24\%). This suggests that case particles may impose an additional segmentation and prediction burden when models process unseen neologism forms.

\subsubsection{Results on Task 2}
From Table \ref{tab:overall_results}, no model exceeds 50 F1 on Task 2. The best-performing model, GPT-5.4, achieves only 49.96 F1, showing that current LLMs remain far from solving this task. Neither parameter scale nor Korean-centric pretraining reliably predicts performance: EXAONE-4.0 (32B) shows little improvement over the much smaller EXAONE-3.5 (7.8B), and both Korean-specialized EXAONE models perform similarly to the general-purpose Qwen3.5.

The difficulty is further increased by the irregularity of truncation, since both the position and unit of truncation (word, syllable, or phoneme) vary unpredictably across items. As shown by GPT-5.4’s incorrect predictions (Table \ref{tab:task2_sequential_scoring_ex} in the Appendix), LLMs often struggle to recover source lexemes from multiple lexical strata. Overall, the results on Task 2 indicate that current LLMs still face substantial difficulty in attaining native-speaker-level understanding of neologisms that emerge dynamically within a language community.

\subsubsection{Results on Task 3}
\paragraph{Semantic Categorization.}
In the semantic categorization Type 1 task, GPT-5.4 and GPT-4.1 achieve 66.80\% and 63.0\%, respectively, whereas EXAONE-4.0, the next best model, reaches only 42.69\%, showing a large gap from the GPT models. This suggests that, regardless of Korean specialization, models with stronger general language understanding are better at comparing semantic similarity and contrast among neologisms. At the same time, even the best results remain in the 60\% range, still indicating clear limitations in this categorization setting.

For Type 2, the best accuracy is lower than in Type 1, which may partly reflect the larger number of target classes (16). However, as shown by the numbers in parentheses in Table \ref{tab:overall_results}, performance improves substantially when usage examples are provided, with gains exceeding 10\% for most models. This result suggests that actively leveraging usage example may improve LLMs’ understanding of neologisms.

\paragraph{Specialized Domain Categorization.}
Overall performance is higher than for semantic categorization, but the gap between the best and worst models exceeds 50\%, making performance differences across models even more pronounced. Although the number of specialized field categories (43) is larger than that of semantic categories, their boundaries are relatively clearer, which likely contributes to the higher performance in Type 1. Still, the gap between the GPT models and the others remains large: the GPT series scores above 80\%, whereas LLaMA-3.1 stays in the 30\% range, possibly reflecting differences in the diversity and domain coverage of the training data.

As with semantic categorization, Type 2 in specialized domain categorization also benefits from the use of usage examples, although the improvement is relatively modest. This is likely because identifying a specialized field requires the model to infer not only the knowledge domain but also its area of use, which cannot be fully inferred from local context alone.

\subsubsection{Results on Task 4}
The Task 4 scores in Table~\ref{tab:overall_results} are obtained through automatic evaluation by Gemini as the judge LLM. GPT-5.4, the best-performing model, achieves 7.49 out of 10 without usage examples and 8.69 with usage examples, indicating very strong performance. In contrast, Solar-10.7B, the lowest-performing model, scores only 3.50 without usage examples and 4.41 even when examples are provided, showing a clear performance gap across models. A detailed breakdown by evaluation criterion is provided in Figure \ref{fig:generation scores} in the Appendix.

\begin{figure}[t]
    \centering
    \includegraphics[width=\columnwidth]{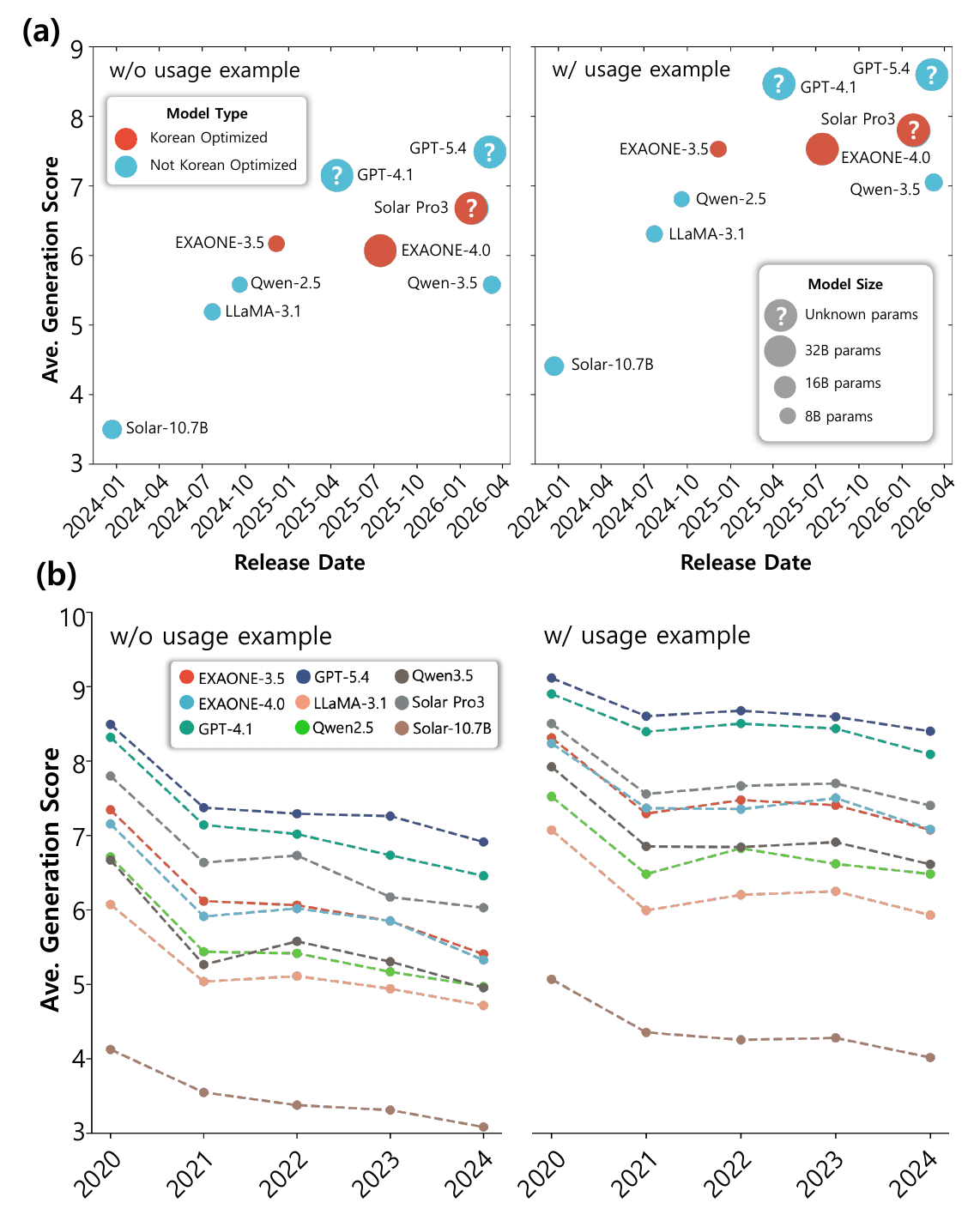}
    \caption{Analysis of LLMs' average scores on Task 4: (a) Performance by model release dates and scales (b) Performance by survey year of neologisms. }
    \label{fig:Task4_scores_distribution}
\end{figure}

Generating a dictionary-style definition and evaluating it against a gold definition requires a model to articulate the meaning of a neologism rather than select among given options. We therefore conduct a more detailed analysis of Task 4, shown in Figure \ref{fig:Task4_scores_distribution}.
Figure \ref{fig:Task4_scores_distribution}(a) shows the relationship between model release date and performance, with marker size indicating model scale. In both the setting with usage examples (left) and the setting without usage examples (right), more recently released models tend to perform better. 
Also, the setting with usage examples shows greater separation across models, suggesting that it is a more informative evaluation setup. Additionally, Korean-specialized models (red markers) generally maintain relatively strong performance regardless of scale or release date.

Figure \ref{fig:Task4_scores_distribution} (b) shows model performance grouped by survey year in which the neologisms first appeared. In the setting without usage examples (left), performance declines clearly for more recent neologisms, and this trend is observed fairly consistently across models. With usage examples (right), overall performance improves, and the decline for more recent neologisms is also reduced. The corresponding year-wise performance distributions for the other tasks are provided in the Appendix.

To validate the automatic evaluation by the judge LLM, we asked a lexicography expert to assess 250 definitions generated by GPT-5.4 and compared the scores. Figure~\ref{fig:task4_human_vs_gemini} in the Appendix shows the distributions of the Gemini and human evaluation scores. The average scores and linear regression results indicate that Gemini tends to assign slightly higher scores than the human evaluator, but the two still show a strong positive correlation (r = 0.708). A detailed analysis is provided in Appendix D.

\section{Discussions}
In this section, we discuss a number of observations arising from the results and consider their implications for understanding and evaluating LLMs’ ability to process Korean neologisms. 

\paragraph{Definition Generation as a Measure of Korean Neologism Understanding.}
Comparing the results across the four tasks suggests that definition generation (Task 4) is a particularly useful measure for evaluating LLMs’ understanding of Korean neologisms. It shows greater performance separation across models than the other tasks and also correlates well with the evaluation results of the remaining tasks (See Figure \ref{fig:Task_correlation_heatmap}). We therefore present definition generation as a complementary and comparatively comprehensive task, since it requires models to articulate meaning rather than select an answer. In the case of Korean, this task also has a practical advantage, since annual neologism dictionaries are published regularly, making it possible to obtain gold definitions without additional large-scale expert annotation. While the reliability of LLM-as-a-judge remains an important issue, the evaluation criteria and prompting strategy proposed in this work help address this issue. This may therefore be considered one of the main practical contributions of our study.

\paragraph{Challenges in LLMs’ Understanding of Korean Neologisms.}
The results across tasks point to three main challenges for LLMs in understanding Korean neologisms. First, time drift and low frequency make neologisms difficult to learn, as shown by the lower performance on more recent items in Figure \ref{fig:Task4_scores_distribution}. Second, the agglutinative structure of Korean introduces additional difficulty, as reflected in the weaker performance on particle-attached forms in Task 1, which is unlikely to be captured by English-centered benchmarks. Third, LLMs still struggle with the formation patterns of Korean neologisms. The below-50\% results for all models in Task 2 suggest that this limitation cannot be addressed simply by scaling up training data alone. 

\paragraph{Using Examples to Improve Neologism Understanding.}
Despite these challenges, the results of Task 3 and Task 4 show substantial performance gains when usage examples are provided (see Table \ref{tab:overall_results}). This suggests that, in real-world settings, LLMs may better understand neologisms by retrieving and leveraging usage examples through retrieval-augmented methods.

\section{Conclusion}
In this paper, we introduced KoNeoBench, the first benchmark for multi-faceted evaluation of LLMs’ understanding of Korean neologisms. Built on a five-year collection of Korean neologisms annotated by lexicography experts, KoNeoBench covers tasks such as source word identification, semantic categorization, and definition generation, while reflecting linguistic properties specific to Korean. By treating neologisms as a distinct evaluation target, it reveals challenges in real-world language use that are not fully captured by existing Korean benchmarks centered on standard vocabulary.

Using KoNeoBench, we proposed an evaluation framework for Korean neologism understanding, analyzed the major sources of difficulty for current LLMs, and showed that usage examples can substantially improve performance. We hope KoNeoBench will serve as a useful step toward evaluation frameworks that better capture ongoing language change in Korean.

\section*{Limitations}
The neologism set used in this study is limited to items observed between 2020 and 2024. 
Earlier neologisms were excluded in light of the fact that recent LLMs are mainly trained on data extending beyond 2023. Neologisms from 2025 are currently being curated and will be incorporated into a future release. 
In addition, semantic neologisms, in which existing words acquire new meanings, are excluded because their first occurrence dates are difficult to determine reliably. The collection and evaluation of semantic neologisms will be addressed in future work.
KoNeoBench is based on neologisms collected from online news articles. While many neologisms first appear in online communities or other informal settings, news media provide a practical source for large-scale collection and verification. Some recent expressions, especially those used in informal or community-specific settings, may be less well represented before they appear in news coverage. Future versions of the benchmark will extend the collection to social media and online communities, with additional procedures for verification and tracking first attestation dates.

\section*{Acknowledgments}
This work was supported by the National Research Foundation of Korea(NRF) grant funded 
by the Korea government(MSIT) (RS-2026-25487705).
This research was supported by the Yonsei University Research Fund of 2026 (2026-22-0116).

\bibliography{custom}

\clearpage
\appendix

\input{appendix}

\end{document}

%% file: appendix.tex
\section{Data Construction}
\label{app:A}

\subsection{Data Collection Process}
\label{app:A.1}
This appendix describes the detailed criteria and yearly statistics of the Korean neologism collection procedure summarized in Section \ref{sec:data_collection}. Our neologism list is based on annual collections compiled from 2020 to 2024 under a consistent methodology. The process consists of four stages: automatic candidate collection, filtering, expert validation, and manual addition. 

\paragraph{Automatic Candidate Collection.}
In this stage, unregistered noun candidates were extracted from the yearly corpus using LRNounExtractorv2, an unsupervised Korean noun extractor. The tool segments sentences into content-word and function-word pairs and extracts candidate single and compound nouns based on frequency and distributional information. Because the extracted set includes not only genuine neologisms but also incomplete fragments, failed analyses, and misspellings, additional filtering is required.

\paragraph{Automatic Filtering.} This stage removes candidates unlikely to be neologisms. We exclude (1) forms containing characters other than Hangul or Roman letters, (2) forms already attested before the target year by comparison with a large inventory of about 330 million word forms, (3) forms listed in the \textit{Standard Korean Language Dictionary}, and (4) inflected or plural forms ending in particles or verbal endings. We also remove candidates that are found, through \textit{Naver News} search, to have been used before the target year.

Because Korean is agglutinative, nouns combine with many particles and predicates appear in many inflected forms. Spacing also does not always match lexical boundaries, and spacing variation is common in actual use. As a result, automatic extraction of unregistered forms alone does not reliably identify neologism candidates. This is especially true for low-frequency items. This limitation is reflected in the collection statistics. As shown in Table \ref{tab:auto_extraction_stats}, automatic extraction produced 1,627,371 candidates from 2020 to 2024, but only 1,269 were finally accepted as neologisms after expert review. 

\begin{table}[t]
\centering
\small
\begin{tabularx}{\columnwidth}{XccX}
\toprule
Year & Candidates & Expert-valid. & Ratio (\%) \\
\midrule
2020 & 409,754 & 237 & 0.06 \\
2021 & 193,905 & 135 & 0.07 \\
2022 & 246,595 & 211 & 0.09 \\
2023 & 322,177 & 282 & 0.09 \\
2024 & 454,940 & 404 & 0.09 \\
\midrule
Total & 1,627,371 & 1,269 & 0.08 \\
\bottomrule
\end{tabularx}
\caption{Yearly statistics of automatically extracted candidates and retained neologisms after expert validation.}
\label{tab:auto_extraction_stats}
\end{table}

\begin{table*}[t] 
\scriptsize 
\setlength{\tabcolsep}{2.5pt} 
\renewcommand{\arraystretch}{1.15} 
\begin{tabularx}{\textwidth}{@{}
>{\centering\arraybackslash}p{0.045\textwidth}
>{\centering\arraybackslash}p{0.055\textwidth}
>{\centering\arraybackslash}p{0.065\textwidth}
>{\centering\arraybackslash}p{0.065\textwidth}
>{\centering\arraybackslash}p{0.055\textwidth}
>{\centering\arraybackslash}p{0.055\textwidth}
>{\centering\arraybackslash}p{0.065\textwidth}
>{\raggedright\arraybackslash}X
>{\raggedright\arraybackslash}X
>{\centering\arraybackslash}p{0.035\textwidth}
>{\raggedright\arraybackslash}p{0.055\textwidth}
@{}}

\toprule
\makecell[c]{\textbf{조사}\\\textbf{연도}} &
\makecell[c]{\textbf{색인}\\\textbf{표제어}} &
\makecell[c]{\textbf{원어}} &
\makecell[c]{\textbf{단어}\\\textbf{형성법}} &
\makecell[c]{\textbf{일상어/}\\\textbf{전문어}} &
\makecell[c]{\textbf{전문}\\\textbf{분야}} &
\makecell[c]{\textbf{의미}\\\textbf{범주}} &
\makecell[c]{\textbf{뜻풀이}} &
\makecell[c]{\textbf{용례}} &
\makecell[c]{\textbf{빈도}} &
\makecell[c]{\textbf{구성}\\\textbf{성분}\\\textbf{분석}} \\
[3pt]

\makecell[c]{\tiny\textbf{(Survey}\\[-1pt]\tiny\textbf{Year)}} &
\makecell[c]{\tiny\textbf{(Head}\\[-1pt]\tiny\textbf{word)}} &
\makecell[c]{\tiny\textbf{(Source}\\[-1pt]\tiny\textbf{form)}} &
\makecell[c]{\tiny\textbf{(Word-}\\[-1pt]\tiny\textbf{formation}\\[-1pt]\tiny\textbf{type)}} &
\makecell[c]{\tiny\textbf{(Domain}\\[-1pt]\tiny\textbf{type)}} &
\makecell[c]{\tiny\textbf{(Specialized}\\[-1pt]\tiny\textbf{Domain)}} &
\makecell[c]{\tiny\textbf{(Semantic}\\[-1pt]\tiny\textbf{category)}} &
\makecell[c]{\tiny\textbf{(Definition)}} &
\makecell[c]{\tiny\textbf{(Usage)}} &
\makecell[c]{\tiny\textbf{(Freq.)}} &
\makecell[c]{\tiny\textbf{(Source}\\[-1pt]\tiny\textbf{words}\\[-1pt]\tiny\textbf{identi}\\[-1pt]\tiny\textbf{-fication}} \\
\midrule

2020 &
먹튜브 &
(먹$\leftarrow$ Youtube) &
혼성 &
일상어 &
-- &
사회생활 &
먹는 방송을 주제로 하는 동영상 공유 서비스 채널을 이르는 말. &
¶한 도시에서 지역 음식점을 발굴하기 위해 해당 지역을 기반으로 하는 \{먹튜브를\} 찾고 있다.  &
89 &
\makecell[c]{\\먹다+\\ 유투브} \\ \hline

2021 &
오하운 &
(오하運) &
축약 &
일상어 &
-- &
삶 &
‘오늘 하루 운동’을 줄여 이르는 말. 매일 일상적으로 하는 운동을 가리키는 말이다. &
¶젊은 세대 사이에는 자존감과 성취감을 높이기 위한 \{오하운이\} 중요한 키워드로 자리를 잡고 있다. &
310 &
\makecell[c]{\\오늘+\\하루+\\운동}\\ \hline

2023 &
자서망 &
(自署網) &
합성 &
전문어 &
정보 통신 &
사회생활 &
경찰 조직 내부에서 사용하는 통신망. 주로 관할 지역 내에서 발생한 긴급 상황을 전파하기 위해 사용한다. &
¶경찰서 내 다양한 통신망 중, 112 신고 전화 등의 긴급 상황은 \{자서망을\} 통해 가장 먼저 전파된다. &
82 &
\makecell[c]{\\--} \\ \hline

2024 &
스키코어 &
(<영>skicore) &
차용 &
전문어 &
복식 &
의생활 &
추운 날씨가 지속되면서, 스키장에서뿐만 아니라 일상생활에서도 입을 수 있는 스키웨어를 착용하는 패션 스타일. &
¶정통 스키 스포츠 브랜드인 OO가 올 겨울 트렌드인 \{스키코어를\} 반영하여 어떤 기상 조건에서도 적응할 수 있는 새로운 컬렉션을 출시했다고 밝혔다. &
74 &
\makecell[c]{\\스키+\\놈코어}
\\

\bottomrule
\end{tabularx}
\caption{Example entries from KoNeoBench.}
\label{tab:dataset_attribute_examples}
\end{table*}

\paragraph{Expert Validation.} Final decisions on neologismhood were made through expert review. In the source data, an item was accepted as a neologism if it (1) referred to a newly emerged object, concept, or phenomenon, (2) expressed a new nuance or attitude despite the existence of a conventional expression, 
or (3) functioned as a phrasal unit with a unified lexical meaning. Proper nouns, expressions restricted to individuals or specific groups, nonce words, and foreign words not established in Korean were excluded. 

\paragraph{Manual Addition.} To address the limitations of automatic extraction, we also conducted manual collection. This step is particularly important for low-frequency items. We relied on pragmatic markers, such as “a neologism called $x$”, “so-called $x$” and similar contextual cues that indicate the use of a new or unfamiliar expression. Such markers have been discussed as useful indicators of both the novelty and the establishment of neologisms \citep{klosa2020considerations,lee2024corpus}. 

\subsection{Data Attributes}
\label{app:A.2}

The KoNeoBench neologism list contains Korean neologisms collected from 2020 to 2024, together with attributes such as survey year, lemma form, source form and word-formation type, semantic category, specialized field, definition, and usage example. The data is released in CSV format, and Table \ref{tab:dataset_attribute_examples} shows sample entries. The attributes are described below.

\begin{itemize}
    \item \textbf{Survey year} indicates when a neologism was first identified within the collection period. Because each annual survey covers news articles published from July 1 of the previous year to June 30 of the target year, a 2020 neologism refers to an item first attested between July 1, 2019 and June 30, 2020. This attribute is used to analyze yearly distributions and temporal changes in model performance.
    \item \textbf{Index form} is the normalized form for indexing and sorting. It removes spacing, hyphens, and other boundary markers from the registered lemma, so that even phrasal expressions and compounds are represented as a single continuous string.
    \item \textbf{Source form} indicates the original elements of a neologism. Native Korean items are given in Hangul, Sino-Korean items in Chinese characters, and loanwords in their original forms with source-language labels. This attribute helps identify whether a neologism is formed from native Korean, Sino-Korean, loanwords, or a mixture of them.
    \item \textbf{Word-formation type} specifies how a neologism was created, including root creation, prefixation, suffixation, compounding, borrowing, abbreviation, and blending. This attribute is particularly important for abbreviation and blending, where the surface form often differs from the underlying constituents.
    \item \textbf{Domain type (general/terminological)} distinguishes everyday neologisms from technical terms used in specific academic, industrial, or social domains. 
    \item \textbf{Specialized domain} specifies the concrete domain of a neologism when it is classified as terminological neologism. The categorization follows the specialized-field taxonomy of `Urimalsaem', which contains 67 fields in total. The terminological neologisms in our dataset are distributed over 43 of these fields. This attribute can be used as the gold label for specialized domain categorization.
    \item \textbf{Semantic category} indicates the semantic domain of the concept denoted by a neologism. It is based on the lexical semantic category system of the \textit{Basic Korean Dictionary}, with additional categories such as health \& medicine and information \& technology to reflect recent neologisms, resulting in a 16-category system. 
    \item \textbf{Definition} is a dictionary-style description of a neologism. It conveys the meaning, function, and usage of the entry, and is used as the gold definition in the definition generation task.
    \item \textbf{Usage Example} shows how a neologism is used in real context. It is based on attested news usage, edited to follow language norms. The headword may be marked with braces, and proper nouns are anonymized.
    \item \textbf{Frequency} indicates how often a neologism appears in the collected data or analysis corpus.
    \item \textbf{Source word identification} is provided only for non-phrasal neologisms and describes their internal morphological structure. It identifies the morphemes or meaningful constituents of a neologism, rather than simply segmenting the surface string. This is especially important for abbreviations and blends, and is directly used in the source word identification task (Task 2).

\end{itemize}

\subsection{Test Set Examples}
\label{app:A.3}
The benchmark dataset described in Appendix~\ref{app:A.2} already includes the answers to Task 2 (source word identification), Task 3-Type 2 (category), and Task 4 (definition). By contrast, Task 1 and Task 3-Type 1 are multiple-choice tasks, so they require candidate options and gold labels for testing and evaluation. This section describes the test sets for Task 1 and Task 3-Type 1.

\subsubsection{Task 1 : Contextual Identification.}
The test set for Task 1 is built from each neologism and its attested usage sentence. We create a cloze question by replacing the target neologism with [MASK]. Each multiple-choice item includes the correct answer, one contextually plausible hard negative, and three random distractors sampled from the benchmark dataset. The hard negative may be conceptually or morphologically similar to the target, semantically contrastive, in a hypernym–hyponym relation, or related in sociocultural background or usage context. Although such hard negative distractors may fit the sentence grammatically or appear partly relevant, they do not fully match the exact referent, semantic focus, or pragmatic nuance of the target neologism.

We further define two input conditions. In the first condition (Type 1), only the target neologism is replaced by [MASK], while the original case particle remains in the sentence. All answer candidates are therefore presented in lemma form. In the second condition (Type 2), both the neologism and the attached particle are replaced by [MASK], and each candidate is presented with an appropriate particle. By comparing these two conditions, we examine whether models rely on morphosyntactic cues such as case particles when selecting the correct answer.

\begin{table}[t]
\centering
\small
\setlength{\tabcolsep}{4pt}
\begin{tabularx}{\columnwidth}{@{}c Y Y@{}}
\toprule
\textbf{Option} & \textbf{Term} & \textbf{Semantic category} \\
\midrule

A 
& 아동양육쿠폰 \newline \textit{(childcare voucher)}
& Politics and administration \\

B 
& \gold{안심 가림막 \newline \textit{(protective partition)}}
& \gold{Health and medicine} \\

C 
& 아베노마스크 \newline \textit{(Abenomask;\scriptsize Abe's mask)}
& Politics and administration \\

D 
& 안보 삼각형 \newline \textit{(security triangle)}
& Politics and administration \\

E 
& 약물 치유 법원 \newline \textit{(drug treatment court)}
& Politics and administration \\

\bottomrule
\end{tabularx}
\caption{Test example from Task 3 (semantic category), Type 1: Five candidates, including the correct answer (shown in blue)}
\label{tab:app-task3-sem-type1}
\end{table}

\begin{table}[t]
\centering
\small
\setlength{\tabcolsep}{4pt}
\begin{tabularx}{\columnwidth}{@{}c Y Y@{}}
\toprule
\textbf{Option} & \textbf{Term} & \textbf{Specialized domain} \\
\midrule

A 
& 안티알고리즘 \newline \textit{(anti-algorithm)}
& Information and communication \\

B 
& 양자비트플랫폼 \newline \textit{(quantum bit platform)}
& Information and communication \\

C 
& \gold{전관용역 \newline \textit{(ex-official service contract)}}
& \gold{General society} \\

D 
& 온디바이스폰 \newline \textit{(on-device phone)}
& Information and communication \\

E 
& 온센서에이아이 \newline \textit{(on-sensor AI)}
& Information and communication \\

\bottomrule
\end{tabularx}
\caption{Test example from Task 3 (specialized domain category), Type 1: Five candidates, including the correct answer (shown in blue).}
\label{tab:app-task3-domain-type1}
\end{table}


\subsubsection{Task 3 (Type 1) : Neologism Categorization.}
Task 3 (Type 1) test set is designed to evaluate whether a model can recognize categorical similarity and difference among multiple neologism candidates. Each question consists of five candidate neologisms. One is the target neologism, and the other four belong to the same category, which is different from that of the target. Thus, the four distractors share a single category, while only the target belongs to a different one. The model is asked to select the item whose category differs from the others, making the target the correct answer.

For semantic categorization, we use the 16 semantic categories assigned to the full dataset. For each target, we randomly choose a category different from its gold label, sample four neologisms from that category, and combine them with the target. The candidate order is then shuffled, and the target serves as the correct answer. Table~\ref{tab:app-task3-sem-type1} shows an example. The same procedure is used for specialized domain categorization, except that only terminological neologisms are included; Table~\ref{tab:app-task3-domain-type1} gives an example. To ensure that four same-category distractors can be constructed reliably, categories with too few candidates are excluded from random sampling.

\section{Evaluation Methods}
\label{app:B}

\subsection{LLM Models for Evaluation}
\label{app:B.1}

\begin{table*}[t]
\centering
\small
\setlength{\tabcolsep}{1pt}
\renewcommand{\arraystretch}{1.1}

\begin{tabularx}{\textwidth}{@{}
>{\raggedright\arraybackslash}p{3.8cm}
>{\raggedright\arraybackslash}p{2.5cm}
>{\centering\arraybackslash}p{1.8cm}
>{\centering\arraybackslash}p{1.5cm}
>{\centering\arraybackslash}p{1.7cm}
>{\raggedright\arraybackslash}X
@{}}
\toprule
\textbf{Model} & \textbf{Provider} & \textbf{Access} & \textbf{Model Size} & \textbf{Release Date} & \textbf{Note} \\
\midrule
\textbf{GPT-4.1} & OpenAI & API & Undisclosed & 2025-04-14 & --  \\
\textbf{GPT-5.4} & OpenAI & API & Undisclosed & 2026-03-05 & -- \\
\textbf{Solar Pro3} & Upstage & API & Undisclosed & 2026-01-26 & Korean-optimized commercial model \\
\textbf{Solar-10.7B}-Instruct & Upstage & Open weights & 10.7B & 2023-12-23 & -- \\
\textbf{EXAONE-3.5}-7.8B-Instruct & LG AI Research & Open weights & 7.8B & 2024-12-06 & Korean--English bilingual instruction tuned model \\
\textbf{EXAONE-4.0}-32B & LG AI Research & Open weights & 32B & 2025-07-15 & Korean--English large language model \\
\textbf{Qwen2.5}-7B-Instruct & Qwen & Open weights & 7B & 2024-09-19 & --\\
\textbf{Qwen3.5}-9B & Qwen & Open weights & 9B & 2026-03-09 & --\\
\textbf{Llama-3.1}-8B-Instruct & Meta & Open weights & 8B & 2024-07-23 & --\\
\bottomrule
\end{tabularx}

\captionof{table}{Overview of the large language models evaluated in this study.}
\label{tab:evaluation_models}
\end{table*}

Table~\ref{tab:evaluation_models} summarizes the large language models evaluated in this study. To compare Korean neologism understanding across different model types, we include both commercial API-based models and instruction-tuned open-weight models. The API-based models are GPT-4.1, GPT-5.4, and Solar Pro3. The open-weight models are Solar-10.7B-Instruct, EXAONE-3.5-7.8B-Instruct, EXAONE-4.0-32B, Qwen2.5-7B-Instruct, Qwen3.5-9B, and Llama-3.1-8B-Instruct.

The models were selected based on three main considerations; (1) we include API-based models to examine how recent commercial LLMs perform on Korean neologism understanding; (2) we include the Solar and EXAONE families to analyze the performance of models specialized for Korean or Korean–English bilingual settings; (3) we include Qwen and Llama as general-purpose multilingual open-weight models for comparison.

Table~\ref{tab:evaluation_models} reports each model’s provider, access type, size, release date, and notable characteristics. API-based models are marked as undisclosed when parameter counts are not public, whereas open-weight models are listed by their reported parameter size. Release date is included only as temporal context and does not necessarily indicate the training-data cutoff. All models are evaluated with the same task prompts and procedure described in the main text.

We used deterministic decoding for the selection and classification tasks by setting the temperature to 0. The answer option order was randomized with a fixed seed of 42. For Qwen3.5, the thinking mode was disabled to prevent intermediate reasoning traces from being included in the output. The maximum output length was set according to the expected output format of each task. Task 1 used 16 output tokens, Task 2 used 2048 output tokens, and Task 3 Type 1 used 256 output tokens. For definition generation (Task 4), we limited the maximum output length to 196 tokens. For the LLM-based evaluation step, we used a \texttt{thinking\_budget} of 1024 and set \texttt{maxOutputTokens} to 4096. We used temperature 0.0 and top-$p$ 0.95. The sampling option \texttt{do\_sample} was set to \texttt{[FALSE]}. To prevent generated definitions from containing Hanja, Chinese, or Japanese characters, we enabled a decoding time forbidden script token ban. In the context-based condition, the usage example was provided with the actual target neologism by replacing the \texttt{[MASK]} placeholder with the corresponding headword before prompting the model.

\subsection{Prompts for LLM execution}
\label{app:B.2}
This appendix describes the prompt templates used for LLM execution in KoNeoBench. 
All prompts were designed to specify the role of the model, the input fields, the task objective, and the required output format. All prompts were written in Korean in order to match the language of the benchmark data and to minimize unintended translation effects. However, English translations are provided only for readability and were not used for model inference.
Placeholders enclosed in braces, such as \texttt{\{term\}}, \texttt{\{ex.\}}, \texttt{\{option\_A\}} and \texttt{\{category\}} were replaced with instance-specific values.

\subsubsection{Prompts for Task 1}
\label{app:prompt-task1}

\begin{figure}[!t]
    \centering
    \setlength{\fboxsep}{4pt}
    \setlength{\fboxrule}{0.5pt}
    \fbox{%
    \begin{minipage}{\dimexpr\columnwidth-2\fboxsep-2\fboxrule\relax}
    \fontsize{9pt}{13pt}\selectfont
    \raggedright
    \setlength{\parskip}{1pt}
    \colorbox{green!60!blue!70!white}{%
    \begin{minipage}{\dimexpr\linewidth-2\fboxsep\relax}
    \raggedright
    \textbf{Task 1 Prompt}
    \end{minipage}
    }
    \noindent
    \colorbox{green!60!blue!7!white}{%
    \begin{minipage}{\dimexpr\linewidth-2\fboxsep\relax}
    \fontsize{9pt}{10pt}\selectfont
    \raggedright
    \textbf{Korean Prompt}
    
    \vspace{2pt}
    \par\textbf{[System]} \par
    당신은 신중한 객관식 문제 풀이 모델입니다.\par
    한국어 문장을 읽고 [MASK]에 가장 적절한 하나의 표제어를 고르십시오.\par
    A, B, C, D, E 중 하나의 대문자만 출력하십시오.\par
    선택 이유를 설명하지 마십시오.\par
    
    \textbf{[User]}\par
    Question (용례):\par
    \{question\} \par
    
    \textbf{Options:}\par
    A. \{option\_A\}\par
    B. \{option\_B\}\par
    C. \{option\_C\}\par
    D. \{option\_D\}\par
    E. \{option\_E\}\par
    
    A, B, C, D, E 중 하나의 글자만 답하십시오.
    
    \end{minipage}
    }
    
    {\setlength{\fboxsep}{4pt}
    \noindent
    \colorbox{gray!6}{%
    \begin{minipage}{\dimexpr\linewidth-2\fboxsep\relax}
    \fontsize{9pt}{10pt}\selectfont
    \raggedright
    
    \textbf{Translation for Readability}
    \par\textbf{[System]} \par
    You are a careful multiple-choice solver.\par
    Read the Korean sentence and choose the single best headword for [MASK].\par
    Return only one uppercase letter among A, B, C, D, E.\par
    Do not explain your choice.\par
    
    \textbf{[User]}\par
    Question (용례):\par
    \{question\} \par

    \textbf{Options:}\par
    A. \{option\_A\}\par
    B. \{option\_B\}\par
    C. \{option\_C\}\par
    D. \{option\_D\}\par
    E. \{option\_E\}\par
    
    Answer with only one letter: A, B, C, D, or E.         
    \end{minipage}
    }
    }
    \end{minipage}
    }

\caption{Prompt template for Task 1}
\label{fig:task1_prompt}
\vspace{-15pt}
\end{figure}

Task 1 evaluates whether a model can select the most appropriate Korean neologism for a masked position in an usage sentence. The prompt provides a Korean sentence containing \texttt{\{MASK\}} and five candidate headwords labeled from \texttt{A} to \texttt{E}. The model is instructed to read the sentence and choose the single candidate that best fits the masked position. Figure~\ref{fig:task1_prompt} explains the prompt in both Korean and English. 
In both variants, the user prompt contains the same fields: the masked usage sentence and the five options. 
The system prompt emphasizes careful multiple-choice solving and explicitly prohibits explanatory text, so that the model output can be directly parsed as a single-choice answer.

\subsubsection{Prompts for Task 2}
\begin{figure*}[t]
\centering

\setlength{\fboxsep}{4pt}
\setlength{\fboxrule}{0.5pt}

\fbox{%
\begin{minipage}{0.94\textwidth}
\small
\raggedright
\renewcommand{\baselinestretch}{0.9}\selectfont
\setlength{\parskip}{1pt}
\colorbox{green!60!blue!70!white}{%
\begin{minipage}{\dimexpr\linewidth-2\fboxsep\relax}
\normalsize
\raggedright
\textbf{Task 2 Prompt}
\end{minipage}
}
\noindent
\colorbox{green!60!blue!7!white}{%
\begin{minipage}{\dimexpr\linewidth-2\fboxsep\relax}
\fontsize{9pt}{13pt}\selectfont
\raggedright

\vspace{2pt}
\noindent
당신은 한국어 신어(neologism)의 구성 성분을 분석하는 언어학 전문가입니다.

\vspace{3pt}

\textbf{[중요]}

이 작업은 일반적인 형태소 분석이 아닙니다.
주어진 신어를 표면적으로 기계 분절하는 것이 아니라,
신어를 이루는 실제 구성 성분을 일정한 규칙에 따라 복원하는 작업입니다.
아래의 분석 지침에 따라 입력된 신어의 구성 성분을 분석하십시오.

\vspace{3pt}

\textbf{[분석 지침]}

\begin{enumerate}[label=\arabic*., leftmargin=1.5em, nosep]
    \item 표면 음절을 임의로 자르지 말고, 실제 조어에 사용된 구성 성분을 찾으십시오.
    \item 숫자, 영어, 외국어, 알파벳 이름 등은 모두 한글 표기로 적으십시오.
    \item 신어 구성 요소에 용언(동사, 형용사)이 포함된 경우, 반드시 기본형 ``-다'' 형태로 복원하십시오.
    \item 축약된 형태는 가능한 경우 원래의 전체 형태로 복원하십시오.
    \begin{itemize}[leftmargin=1.5em, nosep]
        \item 다만, 독립적으로 널리 쓰이거나 사전에 등재된 축약형은 유지할 수 있습니다.
    \end{itemize}
    \item 표준어 규정에 어긋나더라도 실제 언중이 사용하는 비규범적 형태가 구성 성분으로 굳어져 있으면, 표준형으로 고치지 말고 그 형태를 유지하십시오.
    \item 구성 요소가 구(phrase) 단위인 경우, 내부 띄어쓰기는 생략하여 붙여 쓰십시오.
    \item 복수 해석이 가능하더라도 가장 타당한 분석 하나만 제시하십시오.
\end{enumerate}

\vspace{3pt}

\textbf{[작업 방식]}

\begin{enumerate}[label=\arabic*., leftmargin=1.5em, nosep]
    \item 먼저 이 신어가 어떻게 만들어졌는지 \textbf{2--3문장 이내}로 간결하게 분석하십시오.
    \item 각 부분이 어떤 원래 단어에서 왔는지 추론하십시오.
    \item 분석이 끝나면, ``+''로 연결된 구성 성분을 다시 합쳤을 때 원래 신어가 유추 가능한지 검증하십시오.
    \item 검증 후, 마지막 줄에 다음 형식으로 최종 답을 쓰십시오: \par
    \texttt{정답: 구성성분1+구성성분2+...}
\end{enumerate}

\vspace{3pt}

\textbf{[출력 형식]}

\begin{itemize}[leftmargin=1.5em, nosep]
    \item 분석은 2--3문장 이내로 간결하게 작성하십시오.
    \item 마지막 줄의 정답은 ``정답: ''으로 시작해야 합니다.
    \item 정답에는 설명, 괄호, 따옴표, 쉼표, 공백을 넣지 마십시오.
    \item ``+'' 기호로만 구성 성분을 연결하십시오.
\end{itemize}

\vspace{3pt}

\textbf{[예시]}

\begingroup
\leftskip=1em

\textbf{입력: 관태기}\par
분석: ``관태기''는 ``관계''의 앞부분과 ``권태기''의 뒷부분을 합친 혼성어입니다. 인간관계에 권태를 느끼는 시기를 가리키는 말입니다. ``관계''+``권태기''로 합치면 ``관태기''로 축약됩니다.\par
\texttt{정답: 관계+권태기}

\vspace{2pt}

\textbf{입력: 갑분싸}\par
분석: ``갑분싸''는 ``갑자기'', ``분위기'', ``싸하다''의 각 앞글자를 딴 축약어입니다. 용언인 ``싸하다''는 기본형으로 복원해야 합니다. 세 단어를 합치면 원래 표현이 복원됩니다.\par
\texttt{정답: 갑자기+분위기+싸하다}

\vspace{2pt}

\textbf{입력: 망겜}\par
분석: ``망겜''은 ``망하다''의 어간 ``망''과 ``게임''의 축약형 ``겜''을 합친 말입니다. 용언 ``망하다''는 기본형으로 복원합니다. ``망하다''+``게임''으로 분석됩니다.\par
\texttt{정답: 망하다+게임}

\vspace{2pt}

\textbf{입력: 자동봉진}\par
분석: ``자동봉진''은 학교생활기록부의 4대 활동인 ``자율활동'', ``동아리활동'', ``봉사활동'', ``진로활동''의 첫 글자를 딴 축약어입니다. 네 개의 구가 결합된 형태입니다.\par
\texttt{정답: 자율활동+동아리활동+봉사활동+진로활동}

\vspace{2pt}

\textbf{입력: 일코노미}\par
분석: ``일코노미''는 한국어 ``일''(1인)과 영어 economy(이코노미)를 결합한 혼성어입니다. 1인 가구의 소비 경제를 뜻합니다. ``일''+``이코노미''로 분석됩니다.\par
\texttt{정답: 일+이코노미}

\par
\endgroup

\vspace{3pt}

\textbf{[최종 지시]}

\textbf{[중요]}, \textbf{[분석 지침]}, \textbf{[작업 방식]}, \textbf{[출력 형식]}, \textbf{[예시]}를 참고하여 답하시오.

\end{minipage}
}
\end{minipage}
}
\caption{Prompt template for Task 2}
\label{fig:task2_prompt}
\end{figure*}

\begin{figure*}[t]
\centering

\setlength{\fboxsep}{4pt}
\setlength{\fboxrule}{0.5pt}

\fbox{%
\begin{minipage}{0.94\textwidth}
\small
\raggedright
\renewcommand{\baselinestretch}{0.9}\selectfont
\setlength{\parskip}{1pt}
\colorbox{green!60!blue!70!white}{%
\begin{minipage}{\dimexpr\linewidth-2\fboxsep\relax}
\normalsize
\raggedright
\textbf{Task 2 Prompt (Translation for Readability)}
\end{minipage}
}
\noindent
\colorbox{gray!6}{%
\begin{minipage}{\dimexpr\linewidth-2\fboxsep\relax}
\fontsize{9pt}{10.5pt}\selectfont
\raggedright
\noindent
You are a linguistic expert who analyzes the source words of Korean neologisms.

\vspace{3pt}

\textbf{[Important]}

This task is not ordinary morphological analysis.
Rather than mechanically segmenting the given neologism at the surface level,
your task is to reconstruct the actual source words that form the neologism according to a set of rules.
Analyze the source words of the input neologism according to the following guidelines.

\vspace{3pt}

\textbf{[Analysis Guidelines]}

\begin{enumerate}[label=\arabic*., leftmargin=1.5em, nosep]
    \item Do not arbitrarily split surface syllables. Identify the actual source words used in word formation.
    \item Numbers, English words, foreign words, alphabet names, and similar elements should all be written in Korean script.
    \item If a source words includes a predicate, such as a verb or adjective, restore it to its dictionary form ending in ``-다''.
    \item If a component is abbreviated, restore it to its original full form whenever possible.
    \begin{itemize}[leftmargin=1.5em, nosep]
        \item However, abbreviations that are widely used independently or listed in dictionaries may be retained.
    \end{itemize}
    \item If a non-standard form has become established as a source words in actual language use, do not correct it to the standard form; retain the form as used by speakers.
    \item If a source words is a phrase, remove internal spaces and write it as a single unit.
    \item Even if multiple analyses are possible, provide only the most plausible one.
\end{enumerate}

\vspace{3pt}

\textbf{[Procedure]}

\begin{enumerate}[label=\arabic*., leftmargin=1.5em, nosep]
    \item First, briefly analyze how the neologism was formed in no more than \textbf{2--3 sentences}.
    \item Infer which original word each part comes from.
    \item After the analysis, verify whether the original neologism can be inferred by combining the reconstructed source words with ``+''.
    \item After verification, write the final answer on the last line in the following format: \par
    \texttt{Answer: source word1+source word2+...}
\end{enumerate}

\vspace{3pt}

\textbf{[Output Format]}

\begin{itemize}[leftmargin=1.5em, nosep]
    \item Keep the analysis concise, within 2--3 sentences.
    \item The final line must begin with ``Answer: ''.
    \item Do not include explanations, parentheses, quotation marks, commas, or spaces in the final answer.
    \item Connect the source words only with the ``+'' symbol.
\end{itemize}

\vspace{3pt}

\textbf{[Examples]}

\begingroup
\leftskip=1em

\textbf{Input: 관태기}\par
Analysis: ``관태기'' is a blend formed by combining the beginning of ``관계'' and the latter part of ``권태기''. It refers to a period of feeling tired or weary of interpersonal relationships. Combining ``관계'' + ``권태기'' yields the shortened form ``관태기''.\par
\texttt{Answer: 관계+권태기}

\vspace{2pt}

\textbf{Input: 갑분싸}\par
Analysis: ``갑분싸'' is an abbreviation formed from the initial parts of ``갑자기'', ``분위기'', and ``싸하다''. Since ``싸하다'' is a predicate, it should be restored to its dictionary form. The three words together reconstruct the original expression.\par
\texttt{Answer: 갑자기+분위기+싸하다}

\vspace{2pt}

\textbf{Input: 망겜}\par
Analysis: ``망겜'' is formed by combining ``망'', from the verb stem of ``망하다'', and ``겜'', an abbreviated form of ``게임''. The predicate ``망하다'' should be restored to its dictionary form. Therefore, it is analyzed as ``망하다'' + ``게임''.\par
\texttt{Answer: 망하다+게임}

\vspace{2pt}

\textbf{Input: 자동봉진}\par
Analysis: ``자동봉진'' is an abbreviation formed from the first syllables of the four major school record activities: ``자율활동'', ``동아리활동'', ``봉사활동'', and ``진로활동''. It consists of four phrase-level components.\par
\texttt{Answer: 자율활동+동아리활동+봉사활동+진로활동}

\vspace{2pt}

\textbf{Input: 일코노미}\par
Analysis: ``일코노미'' is a blend of the Korean word ``일'', meaning one-person or single-person, and the English word economy, rendered in Korean as ``이코노미''. It refers to the consumer economy of single-person households. It is analyzed as ``일'' + ``이코노미''.\par
\texttt{Answer: 일+이코노미}

\par
\endgroup

\vspace{3pt}

\textbf{[Final Instruction]}

Answer the input by following \textbf{[Important]}, \textbf{[Analysis Guidelines]}, \textbf{[Procedure]}, \textbf{[Output Format]}, and \textbf{[Examples]}.

\end{minipage}
}

\end{minipage}
}
\caption{Prompt template for Task 2 (Translation version)}
\label{fig:task2_prompt_eng}
\end{figure*}

Task 2 evaluates whether a model can reconstruct the underlying components of a Korean neologism. Unlike ordinary morphological analysis, this task does not ask the model to segment the surface form mechanically. Instead, the model is instructed to infer the actual source components involved in the word-formation process and to restore shortened, blended, or transformed elements when appropriate.
Figure~\ref{fig:task2_prompt} presents the prompt of the task2. It provides detailed analysis guidelines. The model should avoid arbitrary syllable-level splitting, convert numbers and foreign-language elements into Korean notation and recover abbreviated forms when possible. The final answer must contain only the reconstructed components connected by plus signs, without parentheses, quotation marks, commas, spaces, or additional explanation.

\subsubsection{Prompts for Task 3}
Task 3 evaluates whether a model can identify the semantic category or specialized domain of Korean neologisms. The task consists of two formats: an odd-one-out format (Task 3 (Type 1)) and a direct category prediction format (Task 3 (Type 2)). 

Task 3(Type 1) evaluates whether a model can identify the neologism whose classification label differs from the other four candidates. Figure~\ref{fig:task3_type1_semantic_prompt} presents the prompt used for the semantic-category setting, and Figure~\ref{fig:task3_type1_domain_prompt} presents the prompt used for the specialized-domain setting. Each prompt presents five Korean neologisms labeled from \texttt{A} to \texttt{E}. 
Among them, four items share the same label under a given classification scheme, while one item belongs to a different label. The model is instructed to select the odd-one-out item based on the actual meaning of each neologism, rather than its surface form or option position. For the semantic-category setting, the prompt provides the complete list of 16 semantic categories. For the specialized-domain setting, the same prompt structure is used with the list of 43 domain labels. Thus, the two versions differ only in the label inventory provided to the model, while the task format, instruction structure, and output constraint remain identical.

\begin{figure*}[t]
\centering

\setlength{\fboxsep}{4pt}
\setlength{\fboxrule}{0.5pt}

\fbox{%
\begin{minipage}{0.94\textwidth}
\small
\raggedright
\renewcommand{\baselinestretch}{0.9}\selectfont
\setlength{\parskip}{1pt}
\colorbox{green!60!blue!70!white}{%
\begin{minipage}{\dimexpr\linewidth-2\fboxsep\relax}
\normalsize
\raggedright
\textbf{Task 3 type 1 (semantic) Prompt}
\end{minipage}
}
\noindent
\colorbox{green!60!blue!7!white}{%
\begin{minipage}{\dimexpr\linewidth-2\fboxsep\relax}
\fontsize{9pt}{13pt}\selectfont
\raggedright
\textbf{=== SYSTEM PROMPT ===}

당신은 한국어 신어의 의미범주를 판별하는 모델입니다.

\vspace{2pt}

이 과제에서 사용하는 의미범주는 다음 16개입니다:

\texttt{\{category\}}

\vspace{2pt}

다섯 개 보기 중 네 개는 같은 의미범주에 속하고, 하나만 다른 의미범주에 속합니다.

각 보기의 실제 의미를 위 의미범주 체계에 따라 판단한 뒤, 나머지 네 개와 의미범주가 다른 하나를 고르십시오.

\vspace{2pt}

출력은 반드시 JSON 형식으로만 작성하십시오.

형식:

\texttt{\{"answer": "A", "reason": "간단한 한 문장 근거"\}}

\vspace{2pt}

규칙:
\begin{itemize}[leftmargin=1.5em, nosep]
    \item \texttt{answer}는 반드시 A, B, C, D, E 중 하나여야 합니다.
    \item \texttt{reason}은 한 문장으로만 작성하십시오.
    \item 자세한 추론 과정은 쓰지 마십시오.
    \item 보기 위치가 아니라 의미범주 차이를 기준으로 판단하십시오.
    \item 주어진 의미범주 목록 밖의 임의 범주를 만들지 마십시오.
\end{itemize}

\vspace{5pt}

\textbf{=== USER PROMPT ===}

아래 다섯 개 신어 중 네 개는 같은 의미범주에 속하고, 하나만 다른 의미범주에 속합니다.

\vspace{2pt}

\texttt{\{options\_text\}}

\vspace{2pt}

위 의미범주 목록을 기준으로, 나머지 네 개와 의미범주가 다른 신어를 고르십시오.
반드시 JSON 형식으로 답하십시오.

\end{minipage}
}

{\setlength{\fboxsep}{4pt}
\noindent
\colorbox{gray!6}{%
\begin{minipage}{\dimexpr\linewidth-2\fboxsep\relax}
\fontsize{9pt}{11pt}\selectfont
\raggedright

\textbf{Translation for Readability}

\textbf{=== SYSTEM PROMPT ===}

You are a model that determines the semantic category of Korean neologisms.

\vspace{2pt}

The semantic categories used in this task are the following 16 categories:

\texttt{\{category\}}

\vspace{2pt}

Among the five options, four belong to the same semantic category, and only one belongs to a different semantic category.
Judge the actual meaning of each option according to the semantic category system above, and select the one whose semantic category differs from the other four.

\vspace{2pt}

You must output only in JSON format.

Format:

\texttt{\{"answer": "A", "reason": "A brief one-sentence justification"\}}

\vspace{2pt}

Rules:
\begin{itemize}[leftmargin=1.5em, nosep]
    \item \texttt{answer} must be one of A, B, C, D, or E.
    \item \texttt{reason} must be written in exactly one sentence.
    \item Do not provide a detailed reasoning process.
    \item Make your decision based on the semantic category difference, not on the position of the options.
    \item Do not create arbitrary categories outside the given semantic category list.
\end{itemize}

\vspace{5pt}

\textbf{=== USER PROMPT ===}

Among the following five Korean neologisms, four belong to the same semantic category, and only one belongs to a different semantic category.

\vspace{2pt}

\texttt{\{options\_text\}}

\vspace{2pt}

Based on the semantic category list above, select the neologism whose semantic category differs from the other four.
You must answer in JSON format.

\end{minipage}
}
}
\end{minipage}
}

\caption{Prompt template for Task 3 (Type 1) (semantic categorization)}
\label{fig:task3_type1_semantic_prompt}
\end{figure*}

\begin{figure*}[t]
\centering

    \setlength{\fboxsep}{4pt}
    \setlength{\fboxrule}{0.5pt}
    
    \fbox{%
    \begin{minipage}{0.94\textwidth}
    \small
    \raggedright
    \renewcommand{\baselinestretch}{0.9}\selectfont
    \setlength{\parskip}{1pt}
    \colorbox{green!60!blue!70!white}{%
    \begin{minipage}{\dimexpr\linewidth-2\fboxsep\relax}
    \normalsize
    \raggedright
    \textbf{Task 3 type 1 (specialized domain) Prompt}
    \end{minipage}
    }
    \noindent
    \colorbox{green!60!blue!7!white}{%
    \begin{minipage}{\dimexpr\linewidth-2\fboxsep\relax}
    \fontsize{9pt}{13pt}\selectfont
    \raggedright
    \textbf{=== SYSTEM PROMPT ===}
    
    당신은 한국어 신어의 전문분야를 판별하는 모델입니다.
    
    \vspace{2pt}
    
    이 과제에서 사용하는 전문분야는 다음 43개입니다:
    
    \texttt{\{category\}}
    
    \vspace{2pt}
    
    다섯 개 보기 중 네 개는 같은 전문분야에 속하고, 하나만 다른 전문분야에 속합니다.
    
    각 보기의 실제 의미를 위 전문분야 체계에 따라 판단한 뒤, 나머지 네 개와 전문분야가 다른 하나를 고르십시오.
    
    \vspace{2pt}
    
    출력은 반드시 JSON 형식으로만 작성하십시오.
    
    형식:
    
    \texttt{\{"answer": "A", "reason": "간단한 한 문장 근거"\}}
    
    \vspace{2pt}
    
    규칙:
    \begin{itemize}[leftmargin=1.5em, nosep]
        \item \texttt{answer}는 반드시 A, B, C, D, E 중 하나여야 합니다.
        \item \texttt{reason}은 한 문장으로만 작성하십시오.
        \item 자세한 추론 과정은 쓰지 마십시오.
        \item 보기 위치가 아니라 전문분야 차이를 기준으로 판단하십시오.
        \item 주어진 전문분야 목록 밖의 임의 범주를 만들지 마십시오.
    \end{itemize}
    
    \vspace{5pt}
    
    \textbf{=== USER PROMPT ===}
    
    아래 다섯 개 신어 중 네 개는 같은 전문분야에 속하고, 하나만 다른 전문분야에 속합니다.
    
    \vspace{2pt}
    
    \texttt{\{options\_text\}}
    
    \vspace{2pt}
    
    위 전문분야 목록을 기준으로, 나머지 네 개와 전문분야가 다른 신어를 고르십시오.
    반드시 JSON 형식으로 답하십시오.
    
    \end{minipage}
    }

    {\setlength{\fboxsep}{4pt}
    \noindent
    \colorbox{gray!6}{%
    \begin{minipage}{\dimexpr\linewidth-2\fboxsep\relax}
    \fontsize{9pt}{11pt}\selectfont
    \raggedright
    
    \textbf{Translation for Readability}
    
    \textbf{=== SYSTEM PROMPT ===}
    
    You are a model that determines the specialized domain category of Korean neologisms.
    
    \vspace{2pt}
    
    The specialized domain categories used in this task are the following 43 categories:
    
    \texttt{\{category\}}
    
    \vspace{2pt}
    
    Among the five options, four belong to the same specialized domain category, and only one belongs to a different specialized domain category.
    Judge the actual meaning of each option according to the specialized domain category system above, and select the one whose specialized domain category differs from the other four.
    
    \vspace{2pt}
    
    You must output only in JSON format.
    
    Format:
    
    \texttt{\{"answer": "A", "reason": "A brief one-sentence justification"\}}
    
    \vspace{2pt}
    
    Rules:
    \begin{itemize}[leftmargin=1.5em, nosep]
        \item \texttt{answer} must be one of A, B, C, D, or E.
        \item \texttt{reason} must be written in exactly one sentence.
        \item Do not provide a detailed reasoning process.
        \item Make your decision based on the specialized domain category difference, not on the position of the options.
        \item Do not create arbitrary categories outside the given specialized domain list.
    \end{itemize}
    
    \vspace{5pt}
    
    \textbf{=== USER PROMPT ===}
    
    Among the following five Korean neologisms, four belong to the same specialized domain category, and only one belongs to a different specialized domain category.
    
    \vspace{2pt}
    
    \texttt{\{options\_text\}}
    
    \vspace{2pt}
    
    Based on the specialized domain category list above, select the neologism whose specialized domain category differs from the other four.
    You must answer in JSON format.
    
    \end{minipage}
    }
    }
    \end{minipage}
    }

\caption{Prompt template for Task 3 (Type 1) (specialized domain categorization)}
\label{fig:task3_type1_domain_prompt}
\end{figure*}

In Task 3 (Type 2), the model is given a single neologism and asked to classify it into the most appropriate category or domain. Figure~\ref{fig:task3_type2_semantic_term_prompt}--Figure~\ref{fig:task3_type2_special_ex_prompt} present the prompt templates used for Task 3 type 2 category prediction. This task evaluates whether LLMs can assign a given Korean neologism to the correct predefined category. We consider two types of labels: 16 semantic categories for general neologisms and 43 specialized domains for terminological neologisms. For each label type, we use two input settings: a `w/o example' setting, where the model receives only the neologism, and a `w/ example' setting, where the model receives both the neologism and its usage example. In all settings, the full label inventory is provided in the prompt, and the model is instructed to output only the corresponding category label, either a letter from A to P for semantic categories or a number from 1 to 43 for specialized domains.
\begin{figure}[t]
\centering
    \centering
    \setlength{\fboxsep}{4pt}
    \setlength{\fboxrule}{0.5pt}
    \fbox{%
    \begin{minipage}{\dimexpr\columnwidth-2\fboxsep-2\fboxrule\relax}
    \fontsize{9pt}{13pt}\selectfont
    \raggedright
    \setlength{\parskip}{1pt}
        
    \colorbox{green!60!blue!70!white}{%
    \begin{minipage}{\dimexpr\linewidth-2\fboxsep\relax}
    \normalsize
    \raggedright
    \textbf{Task 3 type 2 Prompt \\(semantic category, term only) }
    \end{minipage}
    }
    
    \noindent
    \colorbox{green!60!blue!7!white}{%
    \begin{minipage}{\dimexpr\linewidth-2\fboxsep\relax}
    \fontsize{9pt}{13pt}\selectfont
    \raggedright
    
    \textbf{SYSTEM}
    
    한국어 신어의 의미범주를 분류하는 전문가로서, 주어진 신어를 가장 적절한 범주 하나로 분류하시오.
    
    \vspace{5pt}
    
    \textbf{USER}
    
    \textbf{[색인표제어]} \texttt{\{term\}}
    
    \vspace{2pt}
    
    \textbf{[의미범주 목록]}
    
    A: 〔개념〕
    B: 〔경제생활〕
    C: 〔교육〕
    D: 〔동식물〕
    E: 〔문화〕
    F: 〔보건·의학〕
    G: 〔사회생활〕
    H: 〔삶〕
    I: 〔식생활〕
    J: 〔의생활〕
    K: 〔인간〕
    L: 〔자연〕
    M: 〔정보·기술〕
    N: 〔정치와 행정〕
    O: 〔종교〕
    P: 〔주생활〕

    \vspace{2pt}
    
    출력 형식:
    \begin{itemize}[leftmargin=1.5em, nosep]
        \item 반드시 A--P 중 해당하는 하나의 문자만 출력하세요.
        \item 다른 설명은 작성하지 마세요.
    \end{itemize}
    
    \vspace{2pt}
    
    \textbf{정답:}
    
    \end{minipage}
    }
    
    {\setlength{\fboxsep}{4pt}
    \noindent
    \colorbox{gray!6}{%
    \begin{minipage}{\dimexpr\linewidth-2\fboxsep\relax}
    \fontsize{8pt}{11pt}\selectfont
    \raggedright
    
    \textbf{Translation for Readability}
    
    \textbf{SYSTEM}
    
    As an expert in classifying the semantic categories of Korean neologisms, classify the given neologism into the single most appropriate category.
    
    \vspace{5pt}
    
    \textbf{USER}
    
    \textbf{[Headword]} \texttt{\{term\}}
    
    \vspace{2pt}
    
    \textbf{[Semantic category list]}
    
    A: 〔Concept〕
    B: 〔Economic Life〕
    C: 〔Education〕
    D: 〔Animals and Plants〕
    E: 〔Culture〕
    F: 〔Health·Medicine〕
    G: 〔Social Life〕
    H: 〔Life〕
    I: 〔Dietary Life〕
    J: 〔Clothing Life〕
    K: 〔Human〕
    L: 〔Nature〕
    M: 〔Information·Technology〕
    N: 〔Politics and Administration〕
    O: 〔Religion〕
    P: 〔Housing life〕
    
    \vspace{2pt}
    
    Output format:
    \begin{itemize}[leftmargin=1.5em, nosep]
        \item Output only one corresponding letter from A to P.
        \item Do not write any explanation.
    \end{itemize}
    
    \vspace{2pt}
    
    \textbf{Answer:}
    
    \end{minipage}
    }
    }
    
    \end{minipage}
    }

\caption{Prompt template for Task 3 (Type 2) semantic-category with the neologism only.}
\label{fig:task3_type2_semantic_term_prompt}
\end{figure}
\begin{figure}[t]
\centering
    \centering
    \setlength{\fboxsep}{4pt}
    \setlength{\fboxrule}{0.5pt}
    \fbox{%
    \begin{minipage}{\dimexpr\columnwidth-2\fboxsep-2\fboxrule\relax}
    \fontsize{9pt}{13pt}\selectfont
    \raggedright
    \setlength{\parskip}{1pt}
        
    \colorbox{green!60!blue!70!white}{%
    \begin{minipage}{\dimexpr\linewidth-2\fboxsep\relax}
    \normalsize
    \raggedright
    \textbf{Task 3 type 2 Prompt \\(semantic category, with example) }
    \end{minipage}
    }
    
    \noindent
    \colorbox{green!60!blue!7!white}{%
    \begin{minipage}{\dimexpr\linewidth-2\fboxsep\relax}
    \fontsize{9pt}{13pt}\selectfont
    \raggedright
    
    \textbf{SYSTEM}
    
    한국어 신어의 의미범주를 분류하는 전문가로서, 주어진 신어를 가장 적절한 범주 하나로 분류하시오.
    
    \vspace{5pt}
    
    \textbf{USER}
    
    \textbf{[색인표제어]} \texttt{\{term\}}\\
    \textbf{[용례]} \texttt{\{ex.\}}
    
    \vspace{2pt}
    
    \textbf{[의미범주 목록]}
    
    A: 〔개념〕
    B: 〔경제생활〕
    C: 〔교육〕
    D: 〔동식물〕
    E: 〔문화〕
    F: 〔보건·의학〕
    G: 〔사회생활〕
    H: 〔삶〕
    I: 〔식생활〕
    J: 〔의생활〕
    K: 〔인간〕
    L: 〔자연〕
    M: 〔정보·기술〕
    N: 〔정치와 행정〕
    O: 〔종교〕
    P: 〔주생활〕

    \vspace{2pt}
    
    출력 형식:
    \begin{itemize}[leftmargin=1.5em, nosep]
        \item 반드시 A--P 중 해당하는 하나의 문자만 출력하세요.
        \item 다른 설명은 작성하지 마세요.
    \end{itemize}
    
    \vspace{2pt}
    
    \textbf{정답:}
    
    \end{minipage}
    }
    
    {\setlength{\fboxsep}{4pt}
    \noindent
    \colorbox{gray!6}{%
    \begin{minipage}{\dimexpr\linewidth-2\fboxsep\relax}
    \fontsize{8pt}{11pt}\selectfont
    \raggedright
    
    \textbf{Translation for Readability}
    
    \textbf{SYSTEM}
    
    As an expert in classifying the semantic categories of Korean neologisms, classify the given neologism into the single most appropriate category.
    
    \vspace{5pt}
    
    \textbf{USER}
    
    \textbf{[Headword]} \texttt{\{term\}}\\
    \textbf{[Usage]} \texttt{\{ex.\}}
    
    \vspace{2pt}
    
    \textbf{[Semantic category list]}
    
    A: 〔Concept〕
    B: 〔Economic Life〕
    C: 〔Education〕
    D: 〔Animals and Plants〕
    E: 〔Culture〕
    F: 〔Health·Medicine〕
    G: 〔Social Life〕
    H: 〔Life〕
    I: 〔Dietary Life〕
    J: 〔Clothing Life〕
    K: 〔Human〕
    L: 〔Nature〕
    M: 〔Information·Technology〕
    N: 〔Politics and Administration〕
    O: 〔Religion〕
    P: 〔Housing life〕
    
    \vspace{2pt}
    
    Output format:
    \begin{itemize}[leftmargin=1.5em, nosep]
        \item Output only one corresponding letter from A to P.
        \item Do not write any explanation.
    \end{itemize}
    
    \vspace{2pt}
    
    \textbf{Answer:}
    
    \end{minipage}
    }
    }
    
    \end{minipage}
    }

\caption{Prompt template for Task 3 (Type 2) semantic-category with the neologism and Usage.}
\label{fig:task3_type2_semantic_ex_prompt}
\end{figure}

\begin{figure}[t]
\centering
    \centering
    \setlength{\fboxsep}{4pt}
    \setlength{\fboxrule}{0.5pt}
    \fbox{%
    \begin{minipage}{\dimexpr\columnwidth-2\fboxsep-2\fboxrule\relax}
    \fontsize{9pt}{13pt}\selectfont
    \raggedright
    \setlength{\parskip}{1pt}
        
    \colorbox{green!60!blue!70!white}{%
    \begin{minipage}{\dimexpr\linewidth-2\fboxsep\relax}
    \normalsize
    \raggedright
    \textbf{Task 3 type 2 Prompt \\(specialized domain, term only) }
    \end{minipage}
    }
    
    \noindent
    \colorbox{green!60!blue!7!white}{%
    \begin{minipage}{\dimexpr\linewidth-2\fboxsep\relax}
    \fontsize{9pt}{13pt}\selectfont
    \raggedright
    
    \textbf{SYSTEM}
    
    한국어 신어의 전문분야를 분류하는 전문가로서, 주어진 신어를 가장 적절한 분야 하나로 분류하시오.
    
    \vspace{5pt}
    
    \textbf{USER}
    
    \textbf{[색인표제어]} \texttt{\{term\}}
    
    \vspace{2pt}
    
    \textbf{[전문분야 목록]}
    
    1: 『건설』
    2: 『경영』
    3: 『경제』
    4: 『교육』
    5: 『교통』
    6: 『군사』
    7: 『기계』
    8: 『농업』
    9: 『동물』
    10: 『동식물』
    11: 『매체』
    12: 『물리』
    13: 『법률』
    14: 『보건 일반』
    15: 『복식』
    16: 『복지』
    17: 『불교』
    18: 『사회 일반』
    19: 『산업 일반』
    20: 『생명』
    21: 『서비스업』
    22: 『수의』
    23: 『식물』
    24: 『심리』
    25: 『약학』
    26: 『언어』
    27: 『역사』
    28: 『예체능 일반』
    29: 『음악』
    30: 『의학』
    31: 『재료』
    32: 『전기ㆍ전자』
    33: 『정보ㆍ통신』
    34: 『정치』
    35: 『종교 일반』
    36: 『지구』
    37: 『지명』
    38: 『천문』
    39: 『철학』
    40: 『체육』
    41: 『행정』
    42: 『화학』
    43: 『환경』

    \vspace{2pt}
    
    출력 형식:
    \begin{itemize}[leftmargin=1.5em, nosep]
        \item 반드시 목록에 제시된 '번호' 하나만 출력하세요.
        \item 다른 설명은 작성하지 마세요.
    \end{itemize}
    
    \vspace{2pt}
    
    \textbf{정답:}
    
    \end{minipage}
    }
    
    {\setlength{\fboxsep}{4pt}
    \noindent
    \colorbox{gray!6}{%
    \begin{minipage}{\dimexpr\linewidth-2\fboxsep\relax}
    \fontsize{8pt}{11pt}\selectfont
    \raggedright
    
    \textbf{Translation for Readability}
    
    \textbf{SYSTEM}
    
    As an expert in classifying the specialized domain of Korean neologisms, classify the given neologism into the single most appropriate domain.
    
    \vspace{5pt}
    
    \textbf{USER}
    
    \textbf{[Headword]} \texttt{\{term\}}
    
    \vspace{2pt}
    
    \textbf{[Specialized domain list]}
        
    1: [Construction]
    2: [Business Management]
    3: [Economics]
    4: [Education]
    5: [Transportation]
    6: [Military]
    7: [Mechanical Engineering]
    8: [Agriculture]
    9: [Animals]
    10: [Animals and Plants]
    11: [Media]
    12: [Physics]
    13: [Law]
    14: [General Health]
    15: [Clothing and Textiles]
    16: [Welfare]
    17: [Buddhism]
    18: [General Society]
    19: [General Industry]
    20: [Life Science]
    21: [Service Industry]
    22: [Veterinary Medicine]
    23: [Plants]
    24: [Psychology]
    25: [Pharmacy]
    26: [Language]
    27: [History]
    28: [General Arts and Sports]
    29: [Music]
    30: [Medicine]
    31: [Materials]
    32: [Electrical and Electronic Engineering]
    33: [Information and Communication]
    34: [Politics]
    35: [General Religion]
    36: [Earth Science]
    37: [Place Names]
    38: [Astronomy]
    39: [Philosophy]
    40: [Physical Education]
    41: [Public Administration]
    42: [Chemistry]
    43: [Environment]
        
    \vspace{2pt}
    
    Output format:
    \begin{itemize}[leftmargin=1.5em, nosep]
        \item Please make sure to print only one of the 'numbers' presented in the list.
        \item Do not write any explanation.
    \end{itemize}
    
    \vspace{2pt}
    
    \textbf{Answer:}
    
    \end{minipage}
    }
    }
    
    \end{minipage}
    }

\caption{Prompt template for Task 3 (Type 2) specialized domain with the neologism only.}
\label{fig:task3_type2_special_term_prompt}
\end{figure}
\begin{figure}[t]
\centering
    \centering
    \setlength{\fboxsep}{4pt}
    \setlength{\fboxrule}{0.5pt}
    \fbox{%
    \begin{minipage}{\dimexpr\columnwidth-2\fboxsep-2\fboxrule\relax}
    \fontsize{9pt}{13pt}\selectfont
    \raggedright
    \setlength{\parskip}{1pt}
        
    \colorbox{green!60!blue!70!white}{%
    \begin{minipage}{\dimexpr\linewidth-2\fboxsep\relax}
    \normalsize
    \raggedright
    \textbf{Task 3 type 2 Prompt \\(specialized domain, with ex.) }
    \end{minipage}
    }
    
    \noindent
    \colorbox{green!60!blue!7!white}{%
    \begin{minipage}{\dimexpr\linewidth-2\fboxsep\relax}
    \fontsize{9pt}{13pt}\selectfont
    \raggedright
    
    \textbf{SYSTEM}
    
    한국어 신어의 전문분야를 분류하는 전문가로서, 주어진 신어를 가장 적절한 분야 하나로 분류하시오.
    
    \textbf{USER}
    
    \textbf{[색인표제어]} \texttt{\{term\}}\\
    \textbf{[용례]} \texttt{\{ex.\}}
    
    \vspace{2pt}
    
    \textbf{[전문분야 목록]}
    
    1: 『건설』
    2: 『경영』
    3: 『경제』
    4: 『교육』
    5: 『교통』
    6: 『군사』
    7: 『기계』
    8: 『농업』
    9: 『동물』
    10: 『동식물』
    11: 『매체』
    12: 『물리』
    13: 『법률』
    14: 『보건 일반』
    15: 『복식』
    16: 『복지』
    17: 『불교』
    18: 『사회 일반』
    19: 『산업 일반』
    20: 『생명』
    21: 『서비스업』
    22: 『수의』
    23: 『식물』
    24: 『심리』
    25: 『약학』
    26: 『언어』
    27: 『역사』
    28: 『예체능 일반』
    29: 『음악』
    30: 『의학』
    31: 『재료』
    32: 『전기ㆍ전자』
    33: 『정보ㆍ통신』
    34: 『정치』
    35: 『종교 일반』
    36: 『지구』
    37: 『지명』
    38: 『천문』
    39: 『철학』
    40: 『체육』
    41: 『행정』
    42: 『화학』
    43: 『환경』

    \vspace{2pt}
    
    출력 형식:
    \begin{itemize}[leftmargin=1.5em, nosep]
        \item 반드시 목록에 제시된 '번호' 하나만 출력하세요.
        \item 다른 설명은 작성하지 마세요.
    \end{itemize}
    
    \vspace{2pt}
    
    \textbf{정답:}
    
    \end{minipage}
    }
    
    {\setlength{\fboxsep}{4pt}
    \noindent
    \colorbox{gray!6}{%
    \begin{minipage}{\dimexpr\linewidth-2\fboxsep\relax}
    \fontsize{8pt}{11pt}\selectfont
    \raggedright
    
    \textbf{Translation for Readability}
    
    \textbf{SYSTEM}
    
    As an expert in classifying the specialized domain of Korean neologisms, classify the given neologism into the single most appropriate domain.

    \textbf{USER}
    
    \textbf{[Headword]} \texttt{\{term\}}\\
    \textbf{[Usage]} \texttt{\{ex.\}}
    
    \vspace{2pt}
    
    \textbf{[Specialized domain list]}
        
    1: [Construction]
    2: [Business Management]
    3: [Economics]
    4: [Education]
    5: [Transportation]
    6: [Military]
    7: [Mechanical Engineering]
    8: [Agriculture]
    9: [Animals]
    10: [Animals and Plants]
    11: [Media]
    12: [Physics]
    13: [Law]
    14: [General Health]
    15: [Clothing and Textiles]
    16: [Welfare]
    17: [Buddhism]
    18: [General Society]
    19: [General Industry]
    20: [Life Science]
    21: [Service Industry]
    22: [Veterinary Medicine]
    23: [Plants]
    24: [Psychology]
    25: [Pharmacy]
    26: [Language]
    27: [History]
    28: [General Arts and Sports]
    29: [Music]
    30: [Medicine]
    31: [Materials]
    32: [Electrical and Electronic Engineering]
    33: [Information and Communication]
    34: [Politics]
    35: [General Religion]
    36: [Earth Science]
    37: [Place Names]
    38: [Astronomy]
    39: [Philosophy]
    40: [Physical Education]
    41: [Public Administration]
    42: [Chemistry]
    43: [Environment]
        
    \vspace{2pt}
    
    Output format:
    \begin{itemize}[leftmargin=1.5em, nosep]
        \item Please make sure to print only one of the 'numbers' presented in the list.
        \item Do not write any explanation.
    \end{itemize}
    
    \vspace{2pt}
    
    \textbf{Answer:}
    
    \end{minipage}
    }
    }
    
    \end{minipage}
    }

\caption{Prompt template for Task 3 (Type 2) specialized domain with the neologism and example.}
\label{fig:task3_type2_special_ex_prompt}
\end{figure}

\subsubsection{Prompts for Task 4}
Task 4 evaluates whether a model can generate definition for a Korean neologism. We used two prompt variants for this task. 
In the `w/o example' condition, the model receives only the target neologism. Figure~\ref{fig:task4_term_prompt} present the prompt templates for this condition. In the `w/ example' condition, the model receives both the target neologism and one usage example. Figure~\ref{fig:task4_ex_prompt} present the prompt templates for this condition. Both variants include the same definition guidelines. The definition must be direct, concise, non-circular, and semantically clear. 
If pragmatic nuances such as praise or criticism to the meaning of the neologism, the model is instructed to reflect them in the definition when necessary. The required output is exactly one JSON object with two fields: \texttt{term} and \texttt{definition}. The \texttt{term} value must preserve the input headword without modification, and the \texttt{definition} value must contain only one Korean sentence. 
\begin{figure*}[t]
\centering

\setlength{\fboxsep}{4pt}
\setlength{\fboxrule}{0.5pt}

\fbox{%
\begin{minipage}{0.94\textwidth}
\small
\raggedright
\renewcommand{\baselinestretch}{0.9}\selectfont
\setlength{\parskip}{1pt}
\colorbox{green!60!blue!70!white}{%
\begin{minipage}{\dimexpr\linewidth-2\fboxsep\relax}
\normalsize
\raggedright
\textbf{Task 4 Prompt (w/o ex.)}
\end{minipage}
}
\noindent
\colorbox{green!60!blue!7!white}{%
\begin{minipage}{\dimexpr\linewidth-2\fboxsep\relax}
\fontsize{9pt}{13pt}\selectfont
\raggedright
\textbf{[Role]}\\
당신은 한국어 신어 사전 편찬 전문가입니다.\\
입력으로 주어진 표제어(신어, 유행어, 비격식 표현)의 의미를 파악하여 사전식 정의를 작성하십시오.

\textbf{[Task]}\\
표제어만 주어집니다.\\
표제어의 형태, 일반적인 용법, 널리 알려진 의미를 바탕으로 가장 타당한 정의를 작성하십시오.

\textbf{[Definition Guidelines]}\\
1. 정의는 반드시 한국어 한 문장으로 작성합니다.\\
2. 정의는 반드시 사전식 문체로 작성할 것.\\
3. 표제어의 의미를 직접적이고 단정적으로 설명하되, 근거 없는 세부 사항을 임의로 덧붙이지 마십시오.\\
4. 표제어 자체를 그대로 반복하는 순환 정의는 피하십시오.\\
5. 불필요하게 장황하게 쓰지 말고, 핵심 의미가 분명하게 드러나도록 간결하게 작성하십시오.\\
6. 긍정, 부정, 풍자, 비하 등의 뉘앙스가 의미의 핵심이면 정의에 반영하십시오.\\
7. 예문이 없으므로, 특정 상황으로 지나치게 좁히지 말고 보편적으로 통용되는 의미를 중심으로 정의하십시오.\\

\textbf{[Few-shot Examples]}\\
\textbf{[Input]}\\
\{"term": "갓생"\}

\textbf{[Output]}\\
\{"term": "갓생", "definition": "성실하고 모범적으로 자기 관리를 하며 살아가는 삶이나 태도"\}

\textbf{[Output Rules]}\\
-- 반드시 JSON 객체 하나만 출력하십시오.\\
-- 출력 형식은 반드시 다음과 같아야 합니다:\\
  \{"term": "...", "definition": "..."\}\\
-- `term` 값은 입력으로 주어진 표제어를 수정 없이 그대로 사용하십시오.\\
-- `definition` 값은 한국어 한 문장만 포함하십시오.\\
-- 다른 키를 추가하지 마십시오.\\
-- 설명, 주석, 해설, 마크다운 코드블록을 절대 출력하지 마십시오.\\
\textbf{[Final Instruction]}\\
지금부터 주어지는 입력에 대해, 위 규칙을 따른 단일 JSON 객체만 출력하십시오.

\end{minipage}
}

{\setlength{\fboxsep}{4pt}
\noindent
\colorbox{gray!6}{%
\begin{minipage}{\dimexpr\linewidth-2\fboxsep\relax}
\fontsize{8pt}{11pt}\selectfont
\raggedright
\textbf{Translation for Readability}\\
\textbf{[Role]}\\ 
You are an expert lexicographer of Korean neologisms.\\
Identify the meaning of the given headword (neologism, buzzword, or informal expression) and write a dictionary-style definition.

\textbf{[Task]}\\
Only the headword is provided.\\
Based on the form of the headword, its general usage, and its widely known meaning, write the most appropriate definition.

\textbf{[Definition Guidelines]}\\
1. The definition must be written as a single Korean sentence.\\
2. The definition must follow a dictionary-style tone.\\
3. Explain the meaning of the headword directly and assertively, but do not add unsupported details arbitrarily.\\
4. Avoid circular definitions that simply repeat the headword itself.\\
5. Do not write unnecessarily lengthy definitions; keep the definition concise while clearly conveying the core meaning.\\
6. If nuances such as positivity, negativity, satire, or derogation are central to the meaning, reflect them in the definition.\\
7. Since no example sentence is provided, do not narrow the meaning to a specific situation; instead, focus on the meaning that is generally used.

\textbf{[Few-shot Examples]}\\
\textbf{[Input]}\\
\{"term": "갓생"\}

\textbf{[Output]}\\
\{"term": "갓생", "definition": "성실하고 모범적으로 자기 관리를 하며 살아가는 삶이나 태도"\}

\textbf{[Output Rules]}\\
-- Output exactly one JSON object.\\
-- The output format must be exactly as follows:\\
  \{"term": "...", "definition": "..."\}\\
-- The value of `term` must be identical to the input headword without any modification.\\
-- The value of `definition` must contain only one Korean sentence.\\
-- Do not add any other keys.\\
-- Never output explanations, comments, annotations, or Markdown code blocks.\\
\textbf{[Final Instruction]}\\
From now on, for each given input, output only a single JSON object that follows the rules above.

\end{minipage}
}
}
\end{minipage}
}

\caption{Prompt template for Task 4 with the neologism only.}
\label{fig:task4_term_prompt}
\end{figure*}
\begin{figure*}[t]
\centering

\setlength{\fboxsep}{4pt}
\setlength{\fboxrule}{0.5pt}

\fbox{%
\begin{minipage}{0.94\textwidth}
\small
\raggedright
\renewcommand{\baselinestretch}{0.9}\selectfont
\setlength{\parskip}{1pt}
\colorbox{green!60!blue!70!white}{%
\begin{minipage}{\dimexpr\linewidth-2\fboxsep\relax}
\small
\raggedright
\textbf{Task 4 Prompt (w/ ex.)}
\end{minipage}
}
\noindent
\colorbox{green!60!blue!7!white}{%
\begin{minipage}{\dimexpr\linewidth-2\fboxsep\relax}
\fontsize{8pt}{11pt}\selectfont
\raggedright
\textbf{[Role]}\\
당신은 한국어 신어 사전 편찬 전문가입니다. 입력으로 주어진 표제어와 예문을 바탕으로, 해당 신어의 의미를 사전식으로 정확히 정의하십시오.

\textbf{[Task]}\\
입력으로 다음이 주어집니다.\\
- 표제어(term): 한국어 신어 또는 비격식 표현\\
- 예문(example): 해당 표제어가 실제 사용된 문장 1개\\
예문에서 확인되는 용법을 바탕으로 표제어의 핵심 의미를 파악하고, 사전식 문체의 한 문장 정의를 작성하십시오.

\textbf{[Definition Guidelines]}\\
1. 정의는 반드시 한국어 한 문장으로 작성합니다.\\
2. 정의는 반드시 사전식 문체로 작성할 것.\\
3. 예문에 나타난 의미를 충실히 반영하되, 예문의 특정 장면이나 사건에만 묶이지 말고 보편적인 의미로 정리하십시오.\\
4. 다만 예문 하나만으로 확인되지 않는 세부 의미까지 임의로 확장하지 마십시오.\\
5. 예문의 감정적 표현, 과장, 풍자, 비유는 그대로 베끼지 말고, 그 안의 핵심 의미를 객관적으로 정리하십시오.\\
6. 표제어 자체를 그대로 반복하는 순환 정의는 피하십시오.\\
7. 의미상 중요한 뉘앙스(비하, 조롱, 찬양, 자조 등)는 필요할 때 정의에 반영하십시오.\\
8. 정의는 장황하지 않게, 핵심 의미가 분명히 드러나도록 간결하게 작성하십시오.\\

\textbf{[Few-shot Examples]}\\
\textbf{[Input]}\\
\{"term": "갓생", "example": "나 오늘부터 갓생 살 거다."\}

\textbf{[Output]}\\
\{"term": "갓생", "definition": "성실하고 모범적으로 자기 관리를 하며 살아가는 삶이나 태도"\}

\textbf{[Output Rules]}\\
-- 반드시 JSON 객체 하나만 출력하십시오.\\
-- 출력 형식은 반드시 다음과 같아야 합니다:\\
  \{"term": "...", "definition": "..."\}\\
-- `term` 값은 입력으로 주어진 표제어를 수정 없이 그대로 사용하십시오.\\
-- `definition` 값은 한국어 한 문장만 포함하십시오.\\
-- 다른 키를 추가하지 마십시오.\\
-- 설명, 주석, 해설, 마크다운 코드블록을 절대 출력하지 마십시오.\\
\textbf{[Final Instruction]}\\
지금부터 주어지는 입력에 대해, 위 규칙을 따른 단일 JSON 객체만 출력하십시오.

\end{minipage}
}

{\setlength{\fboxsep}{4pt}
\noindent
\colorbox{gray!6}{%
\begin{minipage}{\dimexpr\linewidth-2\fboxsep\relax}
\fontsize{8pt}{9.5pt}\selectfont
\raggedright
\textbf{Translation for Readability}\\
\textbf{[Role]}\\ 
You are an expert lexicographer of Korean neologisms. Based on the given headword and example sentence, accurately define the meaning of the neologism in a dictionary-style format.

\textbf{[Task]}\\
The following inputs are provided.\\
- Headword (term): a Korean neologism or informal expression\\
- Example sentence (example): one sentence in which the headword is actually used\\
Based on the usage observed in the example sentence, identify the core meaning of the headword and write a one-sentence definition in a dictionary-style tone.

\textbf{[Definition Guidelines]}\\
1. The definition must be written as a single Korean sentence.\\
2. The definition must follow a dictionary-style tone.\\
3. Faithfully reflect the meaning shown in the example sentence, but generalize it rather than restricting it to the specific scene or event in the example.\\
4. However, do not arbitrarily expand the definition to include detailed meanings that cannot be confirmed from the single example sentence.\\
5. Do not copy emotional expressions, exaggeration, satire, or figurative wording from the example sentence as they are; instead, objectively summarize the core meaning contained in them.\\
6. Avoid circular definitions that simply repeat the headword itself.\\
7. If semantically important nuances such as derogation, mockery, praise, or self-deprecation are relevant, reflect them in the definition when necessary.\\
8. Do not make the definition verbose; keep it concise while clearly conveying the core meaning.\\

\textbf{[Few-shot Examples]}\\
\textbf{[Input]}\\
\{"term": "갓생", "example": "나 오늘부터 갓생 살 거다."\}

\textbf{[Output]}\\
\{"term": "갓생", "definition": "성실하고 모범적으로 자기 관리를 하며 살아가는 삶이나 태도"\}

\textbf{[Output Rules]}\\
-- Output exactly one JSON object.\\
-- The output format must be exactly as follows:\\
  \{"term": "...", "definition": "..."\}\\
-- The value of `term` must be identical to the input headword without any modification.\\
-- The value of `definition` must contain only one Korean sentence.\\
-- Do not add any other keys.\\
-- Never output explanations, comments, annotations, or Markdown code blocks.\\
\textbf{[Final Instruction]}\\
From now on, for each given input, output only a single JSON object that follows the rules above.
\end{minipage}
}
}
\end{minipage}
}

\caption{Prompt template for Task 4 with the neologism and example.}
\label{fig:task4_ex_prompt}
\end{figure*}

\clearpage
\begin{table*}[t]
\centering
\small
\setlength{\tabcolsep}{3pt}
\renewcommand{\arraystretch}{1.15}
\begin{tabularx}{\textwidth}{@{}
l
>{\raggedright\arraybackslash}p{0.16\textwidth}
>{\raggedright\arraybackslash}p{0.27\textwidth}
>{\raggedright\arraybackslash}p{0.15\textwidth}
>{\raggedright\arraybackslash}p{0.12\textwidth}
>{\raggedright\arraybackslash}p{0.12\textwidth}
@{}}
\toprule
    Case & Gold sequence & Prediction & Aligned & Coverage & Precision \\
    \midrule
    
        Correct
        & 홈$~+~$오마카세
        & 홈$~+~$오마카세 
        & 홈, 오마카세 
        & 2 of 2 = 1.00 
        & 2 of 2 = 1.00 \\

        Partial
        & 홈$~+~$오마카세 
        & 홈
        & 홈
        & 1 of 2 = 0.50  
        & 1 of 1 = 1.00 \\

        Over-generation
        & 홈$~+~$오마카세
        & 홈$~+~$오마카세$~+~$세트 
        & 홈, 오마카세 
        & 2 of 2 = 1.00 
        & 2 of 3 = 0.67 \\

        Wrong
        & 홈$~+~$오마카세 
        & 홈페이지$~+~$마케팅$~+~$가세 
        & -- 
        & 0 of 2 = 0.00 
        & 0 of 3 = 0.00 \\

        Wrong order
        & 홈$~+~$오마카세
        & 오마카세$~+~$홈
        & --
        & 0 of 2 = 0.00
        & 0 of 2 = 0.00 \\
        
    \bottomrule
    \end{tabularx}
\caption{Scoring cases for source word identification (Task 2) of `홈마카세'}
\label{tab:task2_sequential_scoring}
\end{table*}

\begin{table*}[t]
\centering
\small
\setlength{\tabcolsep}{5pt}
\renewcommand{\arraystretch}{1.15}
\begin{tabularx}{\textwidth}{@{}
>{\raggedright\arraybackslash}p{0.14\textwidth}YYYccc
@{}}
\toprule
    Case & Gold sequence & Prediction & Aligned & Coverage & Precision & F1 \\
    \midrule
    
        \makecell[l]{단마토 \\(Danmato;\\Sweet-Tomato)}  
        & \makecell[l]{달다(Sweet)\\$~+~$토마토(Tomato)} 
        & \makecell[l]{달다(Sweet)\\$~+~$토마토(Tomato)} 
        & \makecell[l]{달다(Sweet),\\토마토(Tomato)} 
        & \makecell[l]{~\\2 of 2 = 1.00\\~} 
        & 2 of 2 = 1.00 
        & 1.0\\ 
        \hline \\
        \makecell[l]{헬스닥\\(Hell-sdaq)} 
        & \makecell[l]{헬(Hell)\\$~+~$코스닥(Kosdaq)} 
        & \makecell[l]{헬스(Health)\\$~+~$닥터(Doctor)}
        & --
        & 0 of 2 = 0.00  
        & 0 of 2 = 0.00 
        & 0\\ \\ 
        \hline \\
        \makecell[l]{오하운\\(Ohaun;\\Today's exercise)} 
        & \makecell[l]{오늘(Today)\\$~+~$하루(Day)\\$~+~$운동(Exercise)} 
        & \makecell[l]{오늘(Today)\\$~+~$하체(Lower body)\\$~+~$운동(Exercise)} 
        & \makecell[l]{오늘(Today),\\운동(Exercise)\\~}
        & 2 of 3 = 0.67 
        & 2 of 3 = 0.67 
        & 0.67\\
        \hline \\
        \makecell[l]{스팔로미\\(Th-fallome)}
        & \makecell[l]{스레드(Thread)\\$~+~$팔로미\\~~~~~~(Follow-Me)}
        & \makecell[l]{스포일러(Spoiler)\\$~+~$팔로미\\~~~~~~(Follow-Me)}
        & \makecell[l]{팔로미\\(Follow-Me)}
        & 1 of 2 = 0.50
        & 1 of 2 = 0.50
        & 0.5\\
        
    \bottomrule
    \end{tabularx}
\caption{Examples of GPT-5.4 predictions and scores for Task 2}
\label{tab:task2_sequential_scoring_ex}
\end{table*}

\subsection{Evaluation Criteria for Task 2}
\label{app:B.3}
Since the order of source words is important in Task 2, evaluation is based on source words sequences. Coverage is defined as the proportion of gold source words that are correctly recovered by the model at the correct positions. Precision is defined as the proportion of generated source words that are correctly recovered at the correct positions. Scoring examples are summarized in Table 7. Each example compares a gold source words sequence with a predicted sequence and illustrates how scores are computed based on correctly aligned source words. These examples show how partially correct predictions, order mismatches, and predictions containing unnecessary constituents are evaluated in Task 2. Table 8 further presents a range of example calculations for coverage, precision, and F1 used in evaluating Task 2.

\newpage
\subsection{Evaluation Criteria for Task 4}
\label{app:B.4}
For Task 4, definitions are scored by Gemini in an LLM-as-a-judge setup. Table~\ref{tab:gemini_rubric} presents the scoring criteria defined by lexicography experts. The criteria evaluate three dimensions: semantic adequacy, fluency, and factual reliability.

Semantic adequacy includes three subcriteria. Semantic equivalence checks whether the generated definition preserves the core meaning of the gold definition, even with different wording. Coverage measures whether the key content words of the gold definition are included, capturing possible omissions of essential meaning components. Pragmatic equivalence evaluates whether pragmatic meanings such as evaluative stance, emotion, attitude, disparagement, or mockery are properly reflected when needed.

Fluency includes conciseness, which evaluates whether the definition is brief and free of unnecessary repetition; non-circularity, which checks whether the headword or its synonyms are avoided; and lexicographic convention, which evaluates whether the definition follows dictionary-style metalanguage or a genus–differentia structure. 

Factual reliability assesses whether the generated definition contains hallucinated content, unsupported details, or excessive generalization. The final score is the sum of the subcriterion scores, with a maximum of 10 points. Gemini is instructed to output only the  individual subcriterion scores, not subtotals or the final total. The evaluation prompt is shown in Figure~\ref{fig:task4_judge_prompt}, the output schema in Figure~\ref{fig:task4_eval_schema}, and an English translation of the prompt in Figure~\ref{fig:task4_judge_prompt_eng}.

\begin{figure}[h]
\centering
\setlength{\fboxsep}{7pt}
\setlength{\fboxrule}{0.5pt}

\fbox{%
\begin{minipage}{0.92\columnwidth}
\scriptsize
\raggedright

\textbf{Expected JSON Output Schema for Gemini-based Evaluation}

\vspace{4pt}

\begin{tabular}{@{}p{0.4\textwidth}p{0.53\textwidth}@{}}
\texttt{semantic\_equivalence} & 0 to 2 (Integer). \\
\texttt{coverage} & 0 to 2 (Integer). \\
\texttt{gold\_content\_words} & A list of three objects, each containing \texttt{word} and \texttt{in\_pred}. \\
\texttt{gold\_has\_pragmatic} & Boolean value indicating whether the gold definition contains pragmatic meaning. \\
\texttt{pred\_has\_pragmatic} & Boolean value indicating whether the predicted definition contains pragmatic meaning. \\
\texttt{polarity\_match} & Boolean value indicating whether the pragmatic polarity matches. \\
\texttt{pragmatic\_equivalence} & Either 0 or 2 (Integer). \\
\texttt{conciseness} & Either 0 or 1 (Integer). \\
\texttt{non\_circularity} & Either 0 or 1 (Integer). \\
\texttt{lexicographic\_convention} & Either 0 or 1 (Integer). \\
\texttt{factuality} & Either 0 or 1 (Integer). \\
\texttt{error\_types} & A list of error type strings. \\
\texttt{comment} & A brief natural-language comment. \\
\texttt{confidence} & Numeric confidence score from 0 to 1. \\
\end{tabular}

\end{minipage}
}
\caption{Expected JSON output schema for Gemini-based evaluation of Task~4 definition generation.}
\label{fig:task4_eval_schema}
\end{figure}

\begin{table*}[t]
    \centering
    \small
    \setlength{\tabcolsep}{4pt}
    \renewcommand{\arraystretch}{1.15}
    \begin{tabularx}{\textwidth}{
    >{\raggedright\arraybackslash}p{2.4cm}
    >{\raggedright\arraybackslash}p{3.0cm}lX}
    \toprule
    \textbf{Dimension} & \textbf{Subcriterion} & \textbf{Score} & \textbf{Description} \\
    \midrule
    
    \makecell[tl]{Semantic\\adequacy\\~~(0--6)}
        & Semantic\newline equivalence & 0--2
        & \vspace{-0.8\baselineskip}
          \begin{itemize}[leftmargin=*, nosep, topsep=0pt, partopsep=0pt, itemsep=0pt, parsep=0pt]
            \item Evaluates the semantic equivalence between the gold definition and the predicted definition.\newline(gold definition과 pred definition의 의미적 등가성 평가)
            \item 2 points: core meaning matches
            \item 1 point: core meaning is similar but some details differ
            \item 0 points: core meaning does not match
          \end{itemize} \\
        
        & Coverage & 0--2
        & \vspace{-0.8\baselineskip}
            \begin{itemize}[leftmargin=*, nosep, topsep=0pt, partopsep=0pt, itemsep=0pt, parsep=0pt]
            \item Extracts three key content words from the gold definition and evaluates how many of them are covered by the predicted definition.\newline(gold definition에서 핵심 내용어 3개를 추출하고, pred definition이 이를 얼마나 포함하는지 평가)
            \item 2 points: all three are covered
            \item 1 point: one or two are covered
            \item 0 points: none are covered
          \end{itemize} \\
        
        & Pragmatic\newline equivalence & 0 or 2
        & \vspace{-0.8\baselineskip}
          \begin{itemize}[leftmargin=*, nosep, topsep=0pt, partopsep=0pt, itemsep=0pt, parsep=0pt]
            \item Evaluates whether the pragmatic meaning\textsuperscript{*} of the gold definition and the predicted definition is equivalent. \newline(gold와 pred의 화용적 의미가 일치하는지 평가)
            \item 2 points: both the gold and predicted definitions contain pragmatic meaning with consistent polarity, or neither definition contains pragmatic meaning
            \item 0 points: the presence or polarity of pragmatic meaning is inconsistent
          \end{itemize} \\
    
    \midrule
    
    \makecell[tl]{Fluency\\~~(0--3)}
    & Conciseness & 0--1 
    & \vspace{-0.8\baselineskip}
      \begin{itemize}[leftmargin=*, nosep, topsep=0pt, partopsep=0pt, itemsep=0pt, parsep=0pt]
        \item Evaluates whether the predicted definition avoids unnecessary modifiers or repetitive expressions.\newline(pred definition에 불필요한 수식이나 반복 표현이 없는지 평가)
        \item 1 point: concise
        \item 0 points: contains unnecessary modifiers or repetition
      \end{itemize} \\
    
    & Non-circularity & 0--1
    & \vspace{-0.8\baselineskip}
      \begin{itemize}[leftmargin=*, nosep, topsep=0pt, partopsep=0pt, itemsep=0pt, parsep=0pt]
         \item Evaluates whether the predicted definition avoids circularly defining the target word using the target word itself or its close synonyms \newline(pred definition이 표제어나 표제어의 유의어를 사용하여 순환적으로 정의하지 않는지 평가)
        \item 1 point: non-circular definition
        \item 0 points: circular definition
      \end{itemize} \\
    
    & Lexicographic convention & 0--1
    & \vspace{-0.8\baselineskip}
      \begin{itemize}[leftmargin=*, nosep, topsep=0pt, partopsep=0pt, itemsep=0pt, parsep=0pt]
        \item Evaluates whether the predicted definition appropriately follows lexicographic conventions, such as the use of definitional metalanguage or a genus--differentia structure.\newline(pred definition이 사전 정의문 형식의 메타언어 또는 유개념--종차 구조를 적절히 따르는지 평가)
        \item 1 point: follows lexicographic conventions
        \item 0 points: does not follow lexicographic conventions.
      \end{itemize} \\
    
    \midrule
    
    \makecell[tl]{Factuality\\~~(0--1)}
    & Factuality & 0--1
    & \vspace{-0.8\baselineskip}
      \begin{itemize}[leftmargin=*, nosep, topsep=0pt, partopsep=0pt, itemsep=0pt, parsep=0pt]
        \item Evaluates whether the predicted definition contains hallucinations, unsupported specific claims, or overgeneralizations.\newline (pred definition에 hallucination, 근거 없는 구체적 진술, 과도한 일반화가 있는지 평가)
        \item 1 point: no major issue
        \item 0 points: at least one clear and significant issue is present
      \end{itemize} \\
    
    \bottomrule
    \end{tabularx}
    
    \vspace{2pt}
    \begin{minipage}{\textwidth}
    \footnotesize
    \raggedright
    \textit{Note.} \textsuperscript{*} Pragmatic meaning refers to evaluative or affective meanings associated with the defined term, including attitudes such as support, encouragement, derogation, hatred, or mockery. The total score is 10 points and is computed post hoc as the sum of the subcriterion scores.
    \end{minipage}
    \caption{Gemini-based evaluation rubric for Task 4 definition generation.}
    \label{tab:gemini_rubric}
\end{table*}

\begin{figure*}[t]
    \centering
    
    \setlength{\fboxsep}{6pt}   
    \setlength{\fboxrule}{0.5pt} 
    
    \fcolorbox{black!70}{green!60!blue!7!white}{%
    \begin{minipage}{0.94\textwidth}
    \small
    \raggedright 
    \textbf{Prompt for Task 4 Definition Judgement}
    \vspace{4pt}
    
    당신은 한국어 신어 정의 생성 결과를 엄격하게 채점하는 평가자다.\newline
    주어진 gold definition(정답 정의)와 pred definition(모델 생성 정의)을 비교하여,
    example(용례)을 문맥 정보로 참고해 채점하라. 반드시 아래 A--C의 3개 기준만 사용하고, 모든 점수는 명시된 범위 내의 정수(integer)여야 한다. (음수 불가)

\textbf{A) Semantic Adequacy}

\vspace{2pt}

\begingroup
\leftskip=1em

\textbf{A-1) \textbf{semantic\_equivalence}: 0--2 points}\par
-- gold definition과 predicted definition의 의미적 등가성 평가 \newline
-- 표현이 달라도 의미가 일치하면 높은 점수 부여 (패러프레이즈 허용)

\begin{itemize}[leftmargin=3.0em, nosep]
    \item 2 점 핵심 의미 일치 $|$ 1 점 : 핵심 의미는 일치하나, 세부 사항 불일치 $|$ 0점 : 핵심 의미 불일치
\end{itemize}

\vspace{2pt}

\textbf{A-2) \textbf{coverage}: 0--2 points}\par
-- gold definition에서 핵심적 내용어(content word) 3개 추출 후, pred definition이 이를 얼마나 포함하는지 평가\newline
-- 한국어 동사, 형용사의 활용형은 동일한 내용어임\newline
-- 기능어(Function word)나 불용어(Stopwords) 제외\newline
-- gold definition에서 추출한 내용어와 각 내용어가 pred에 포함되는지 여부를  gold\_content\_words 필드에 출력

\begin{itemize}[leftmargin=3.0em, nosep]
    \item 2 점: 3개 모두 포함 $|$ 1 점: 1~2개 포함 $|$ 0 점: 미포함
\end{itemize}
-- 추가 출력 필드: gold\_content\_words: \text{[\{“word”: “…”, “in\_pred”: true/false\}, …]}  
\vspace{2pt}

\textbf{A-3) \textbf{pragmatic\_equivalence}: 0점 또는 2점}\par
-- gold의 화용적 의미와 pred의 화용적 의미가 일치하는지 평가\newline
    ※ 화용적 의미: 정의문 기술 대상에 대한 평가적 의미(긍정/부정)나 화자의 감정 혹은 태도(지지/응원, 혐오, 비하, 조롱 등) 등. 화용적 의미가 일치한다는 것은 gold와 pred의 극성(polarity)이 일치하는지를 의미한다. 즉, gold가 긍정이며 pred도 긍정, gold가 부정이면 pred도 부정이어야 한다. gold와 pred의 극성이 반대(예: 긍정 ↔ 부정)라면 불일치로 본다.

    다음 다섯 가지 경우를 모두 채점 대상으로 함
    \begin{enumerate}[label=\arabic*), leftmargin=3.0em, nosep]
    \item gold에 화용적 의미 포함 + pred에 화용적 의미 포함 + 극성 일치 = 2점
    \item gold에 화용적 의미 포함 + pred에 화용적 의미 포함 + 극성 불일치 = 0점
    \item gold에 화용적 의미 포함 + pred에 화용적 의미 미포함 = 0점
    \item gold에 화용적 의미 미포함 + pred에 화용적 의미 포함 = 0점
    \item gold에 화용적 의미 미포함 + pred에 화용적 의미 미포함 = 2점
    \end{enumerate}

-- 추가 출력 필드(true/false): gold\_has\_pragmatic, pred\_has\_pragmatic, polarity\_match
\par
\endgroup

\textbf{B) fluency : 0--3점} \newline
-- 정의문의 형식적 적합성 평가 \par
-- 아래 3개 하위 기준을 각각 0점 또는 1점으로 채점
\vspace{2pt}

\begingroup
\leftskip=1em

\textbf{B-1) \textbf{conciseness} (간결성): 0점 또는 1점}\par
-- gold definition과 비교하여, pred definition에 불필요한 수식이나 반복 표현이 없는지 평가

\begin{itemize}[leftmargin=3.0em, nosep]
    \item 1점: 불필요한 수식이나 반복 표현 없이 간결함 $|$ 0점: 불필요한 수식이나 반복 표현이 포함됨
\end{itemize}

\vspace{2pt}

\textbf{B-2) \textbf{non-circularity}(비순환성): 0점 또는 1점}\par
-- pred definition에 해당 표제어나, 해당 표제어의 유의어 등을 사용한 순환적 정의가 없는지 평가

\begin{itemize}[leftmargin=3.0em, nosep]
    \item 1점: 비순환적 정의문 $|$ 0점: 순환적 정의문
\end{itemize}

\vspace{2pt}

\textbf{B-3) \textbf{lexicographic convention} (사전학적 관습 준수): 0점 또는 1점}\par
-- pred definition에 사전 정의문 형식의 메타언어(예: --을 이르는 말 등)나 유개념--종차 구조(예: 소확행 - [소소하지만 확실한](종차) + [행복](유개념))를 적절하게 사용하는지 평가 

\begin{itemize}[leftmargin=3.0em, nosep]
    \item 1점: 사전학적 관습 준수 $|$ 0점: 사전학적 관습 미준수
\end{itemize}

\par
\endgroup

\textbf{C) factuality: 0점 또는 1점} \newline
-- hallucination, 근거 없는 구체적 진술, 과도한 일반화 여부 평가
\begin{itemize}[leftmargin=1.5em, nosep]
    \item 1점: 중대한 hallucination/근거 없는 구체화/과도한 일반화 없음
    \item 0점: 위 문제 중 하나라도 크고 명확하게 존재함
\end{itemize}

    중요 규칙:
\begin{enumerate}[label=\arabic*), leftmargin=3.0em, nosep]
    \item 모든 점수는 명시된 범위 내의 정수여야 한다. (음수 불가)
    \item 출력은 각 하위 항목 점수만 포함하며, 소계(semantic\_adequacy, fluency)와 총점(total)은 출력하지 마라. 합산은 사후 처리한다.
    \item semantic\_equivalence가 맞으면 표현이 달라도 감점하지 마라.
    \item example은 의미를 해석하는 문맥으로만 사용하라. gold를 넘어서는 새로운 의미를 추가하는 근거로 사용하지 마라.
    \item pred가 너무 짧더라도 핵심 의미를 담으면 점수를 줄 수 있다.
    \item pred가 장황하더라도 근거 없는 내용이 들어가면 factuality를 감점하라.
    \item 출력은 반드시 JSON 객체 하나만 반환하라.
    \item markdown, 코드블록, 설명문 없이 아래의 스키마로 JSON만 출력하라.
    \end{enumerate}        
    \end{minipage}
}   
\caption{Prompt template for Task 4 definition generation.}
\label{fig:task4_judge_prompt}
\end{figure*}
\begin{figure*}[t]
    \centering
    
    \setlength{\fboxsep}{6pt}   
    \setlength{\fboxrule}{0.5pt} 
    
    \fcolorbox{black!70}{gray!6}{%
    \begin{minipage}{0.94\textwidth}
    \small
    \raggedright 
    \fontsize{8pt}{9pt}\selectfont
    \textbf{Prompt for Task 4 Definition Judgement(For readability)}
    \vspace{4pt}
    
    You are a strict evaluator of Korean neologism definition generation.\newline
    Compare the given gold definition and predicted definition, using the example only as contextual information. 
    Use only the following three criteria A--C, and all scores must be integers within the specified ranges. Negative scores are not allowed.

\textbf{A) Semantic Adequacy}

\vspace{2pt}

\begingroup
\leftskip=1em

\textbf{A-1) \textbf{semantic\_equivalence}: 0--2 points}\par
-- Evaluate the semantic equivalence between the gold definition and the predicted definition.\newline
-- Assign a high score when the meanings match, even if the wording differs. Paraphrases are allowed.

\begin{itemize}[leftmargin=3.0em, nosep]
    \item 2 points: core meaning matches \textbar{} 1 point: core meaning matches partially but some details differ \textbar{} 0 points: core meaning does not match
\end{itemize}

\vspace{2pt}

\textbf{A-2) \textbf{coverage}: 0--2 points}\par
-- Extract three key content words from the gold definition and evaluate how many of them are covered by the predicted definition.\newline
-- Inflected forms of Korean verbs and adjectives should be treated as the same content word.\newline
-- Function words and stopwords should be excluded.\newline
-- Output the content words extracted from the gold definition and whether each word is included in the predicted definition in the \texttt{gold\_content\_words} field.

\begin{itemize}[leftmargin=3.0em, nosep]
    \item 2 points: all three words are included \textbar{} 1 point: one or two words are included \textbar{} 0 points: none are included
\end{itemize}
-- Additional output field: \texttt{gold\_content\_words}: \texttt{[\{"word": "...", "in\_pred": true/false\}, ...]}  

\vspace{2pt}

\textbf{A-3) \textbf{pragmatic\_equivalence}: 0 or 2 points}\par
-- Evaluate whether the pragmatic meaning of the gold definition and the predicted definition is equivalent.\newline
    Pragmatic meaning refers to evaluative meaning toward the defined term, such as positive or negative evaluation, or the speaker's emotion or attitude, including support, encouragement, hatred, derogation, or mockery. Pragmatic equivalence means that the polarity of the gold and predicted definitions is consistent. That is, if the gold definition is positive, the predicted definition should also be positive; if the gold definition is negative, the predicted definition should also be negative. If the polarity is opposite, such as positive versus negative, it should be considered inconsistent.

    Score all of the following five cases:
    \begin{enumerate}[label=\arabic*), leftmargin=3.0em, nosep]
    \item Gold contains pragmatic meaning + prediction contains pragmatic meaning + polarity matches = 2 points
    \item Gold contains pragmatic meaning + prediction contains pragmatic meaning + polarity does not match = 0 points
    \item Gold contains pragmatic meaning + prediction does not contain pragmatic meaning = 0 points
    \item Gold does not contain pragmatic meaning + prediction contains pragmatic meaning = 0 points
    \item Gold does not contain pragmatic meaning + prediction does not contain pragmatic meaning = 2 points
    \end{enumerate}

-- Additional output fields true/false: \texttt{gold\_has\_pragmatic}, \texttt{pred\_has\_pragmatic}, \texttt{polarity\_match}
\par
\endgroup

\textbf{B) Fluency: 0--3 points} \newline
-- Evaluate the formal appropriateness of the definition.\par
-- Score each of the following three subcriteria as either 0 or 1 point.
\vspace{2pt}

\begingroup
\leftskip=1em

\textbf{B-1) \textbf{conciseness}: 0 or 1 point}\par
-- Evaluate whether the predicted definition avoids unnecessary modifiers or repetitive expressions, compared with the gold definition.

\begin{itemize}[leftmargin=3.0em, nosep]
    \item 1 point: concise without unnecessary modifiers or repetition \textbar{} 0 points: contains unnecessary modifiers or repetitive expressions
\end{itemize}

\vspace{2pt}

\textbf{B-2) \textbf{non-circularity}: 0 or 1 point}\par
-- Evaluate whether the predicted definition avoids circularly defining the target word using the target word itself or its synonyms.

\begin{itemize}[leftmargin=3.0em, nosep]
    \item 1 point: non-circular definition \textbar{} 0 points: circular definition
\end{itemize}

\vspace{2pt}

\textbf{B-3) \textbf{lexicographic convention}: 0 or 1 point}\par
-- Evaluate whether the predicted definition appropriately follows lexicographic conventions, such as using definitional metalanguage or a genus--differentia structure. For example, in ``small but certain happiness,'' ``small but certain'' functions as the differentia and ``happiness'' as the genus.

\begin{itemize}[leftmargin=3.0em, nosep]
    \item 1 point: follows lexicographic conventions \textbar{} 0 points: does not follow lexicographic conventions
\end{itemize}

\par
\endgroup

\textbf{C) Factuality: 0 or 1 point} \newline
-- Evaluate whether the predicted definition contains hallucinations, unsupported specific claims, or overgeneralizations.
\begin{itemize}[leftmargin=1.5em, nosep]
    \item 1 point: no major hallucination, unsupported specificity, or overgeneralization
    \item 0 points: at least one clear and significant issue of the above types is present
\end{itemize}

    Important rules:
\begin{enumerate}[label=\arabic*), leftmargin=3.0em, nosep]
    \item All scores must be integers within the specified ranges. Negative scores are not allowed.
    \item The output should include only the subcriterion scores. Do not output subtotal scores such as \texttt{semantic\_adequacy} or \texttt{fluency}, or the final \texttt{total} score. These scores will be computed post hoc.
    \item If \texttt{semantic\_equivalence} is satisfied, do not penalize differences in wording.
    \item Use the example only as contextual information for interpreting the meaning. Do not use it as evidence to add new meanings beyond the gold definition.
    \item Even if the predicted definition is very short, it may receive credit if it captures the core meaning.
    \item Even if the predicted definition is verbose, penalize factuality if it contains unsupported content.
    \item The output must be exactly one JSON object.
    \item Return only JSON following the schema below, without markdown, code blocks, or explanatory text.
    \end{enumerate}        
    \end{minipage}
}   
\caption{Prompt template for Task 4 definition generation (Eng).}
\label{fig:task4_judge_prompt_eng}
\end{figure*}

\begin{figure*}[h]
\centering
\includegraphics[width=0.8\textwidth]{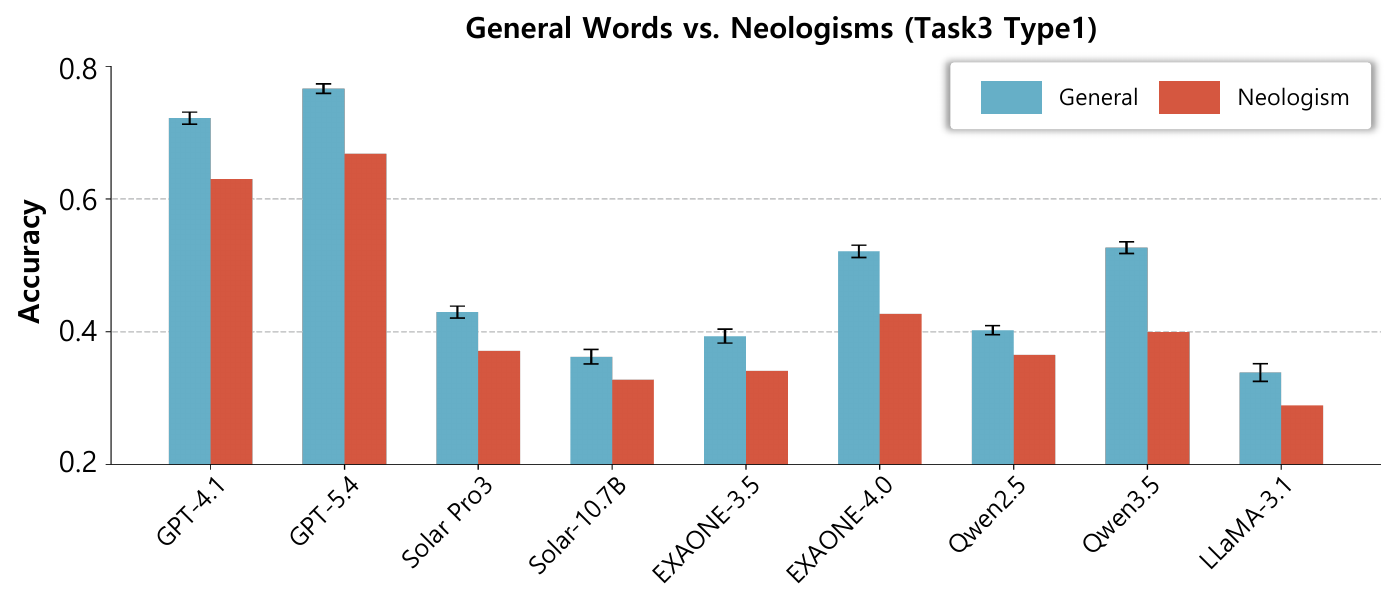} 
\caption{Semantic categorization (Task 3-type 1) accuracy: general words vs. neologisms
}
\label{fig:general_neologism_accuracy_all}
\end{figure*}

\section{Analysis on Evaluation Results}
\label{app:C}

\subsection{Semantic Categorization: General vs. Neologism}
\label{app:C.2}

This section describes how the general-word results in Figure~\ref{fig:fig1} were obtained. Since the performance for neologism is reported using Task 3 (Type 1), we evaluated the performance for general words under the same setup. 
We first selected 16,738 nouns from the headword list of the \textit{Basic Korean Dictionary} that satisfied predefined conditions. From this set, we randomly sampled 1,785 items (the same number as the neologisms) and constructed test sets following the same procedure as Task 3 (Type 1). This experiment was repeated 10 times to obtain the average semantic categorization performance on general words. Results for three representative LLMs are shown in Figure~\ref{fig:fig1}, while Figure~\ref{fig:general_neologism_accuracy_all} reports all nine models. One caveat is that the semantic categories represented in the general-word and neologism sets may vary depending on random sampling. To account for this variation, Figure~\ref{fig:general_neologism_accuracy_all} also reports standard deviations.

\subsection{Overall Performances across Tasks}
\begin{figure*}[ht]
    \centering
    \includegraphics[width=\textwidth]{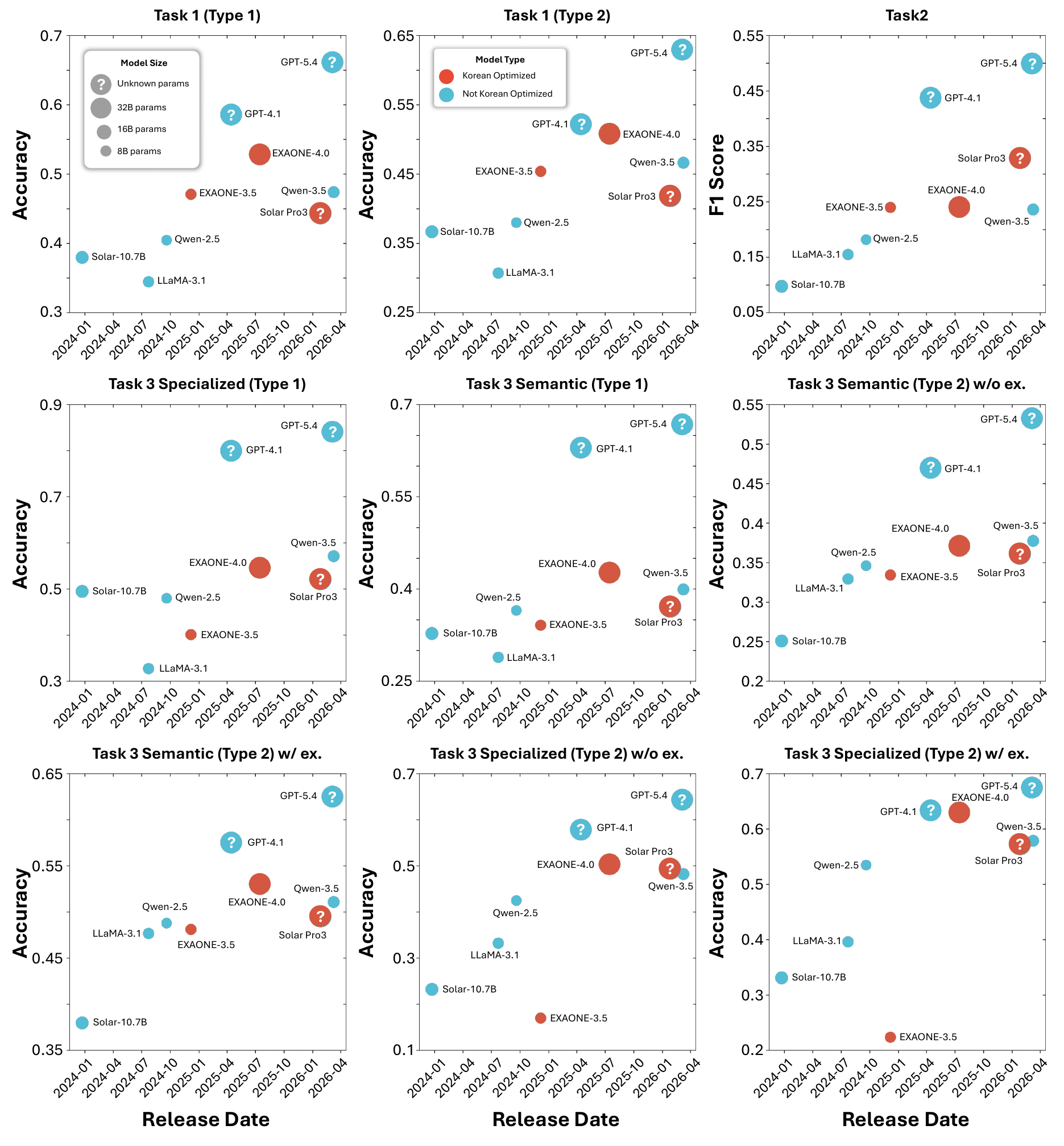}
    \caption{Task-wise performances by model release dates and scales }
    \label{fig:Task_wise_scatter_plot}
\end{figure*}

We further examine whether the sub-tasks in KoNeoBench show common performance trends across models and what task-specific difficulties they reveal. For this analysis, we select one representative setting from each task group: Task 1 (Type 1) for contextual identification, Task 2 (F1) for source word identification, Task 3 (Semantic, Type 1) and Task 3 (Specialized Domain, Type 1) for category-based discrimination, and Task 4 with usage examples for definition generation. Figure~\ref{fig:Task_wise_scatter_plot} and ~\ref{fig:Task4_scores_distribution}(a) shows performance distributions by model release date and scale. Overall, more recent and larger models tend to perform better across tasks. The GPT series is consistently among the strongest, especially in Task 1, Task 3, and Task 4, suggesting relatively strong ability in contextual identification, categorical classification, and definition generation for Korean neologisms.

The most notable difference in difficulty appears in Task 2. Its F1 scores are generally lower than those of the multiple-choice and classification tasks, and model gaps are also larger. This suggests that roughly understanding a neologism and identifying its source words are different challenges. While Task 1 and Task 3 allow models to rely on contextual or categorical cues, Task 2 requires direct identification of source words that are not explicitly present in the surface form.

\begin{figure*}[ht]
    \centering
    \includegraphics[width=\textwidth]{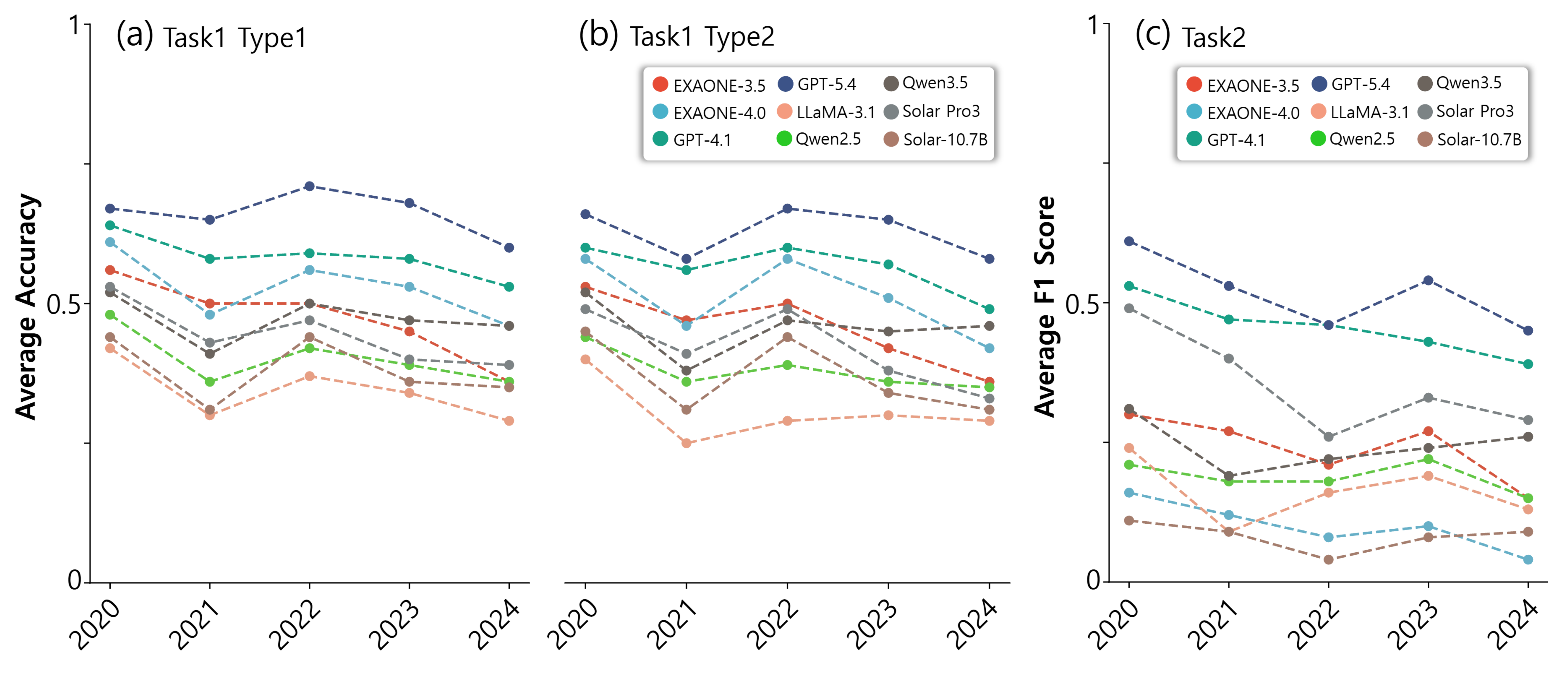}
    \caption{Year-wise performance distributions for Task 1 and Task 2 }
    \label{fig:year-f1}
\end{figure*}

\begin{figure*}[ht]
    \centering
    \includegraphics[
    width=0.72\textwidth,
    height=0.30\textheight,
    keepaspectratio
]{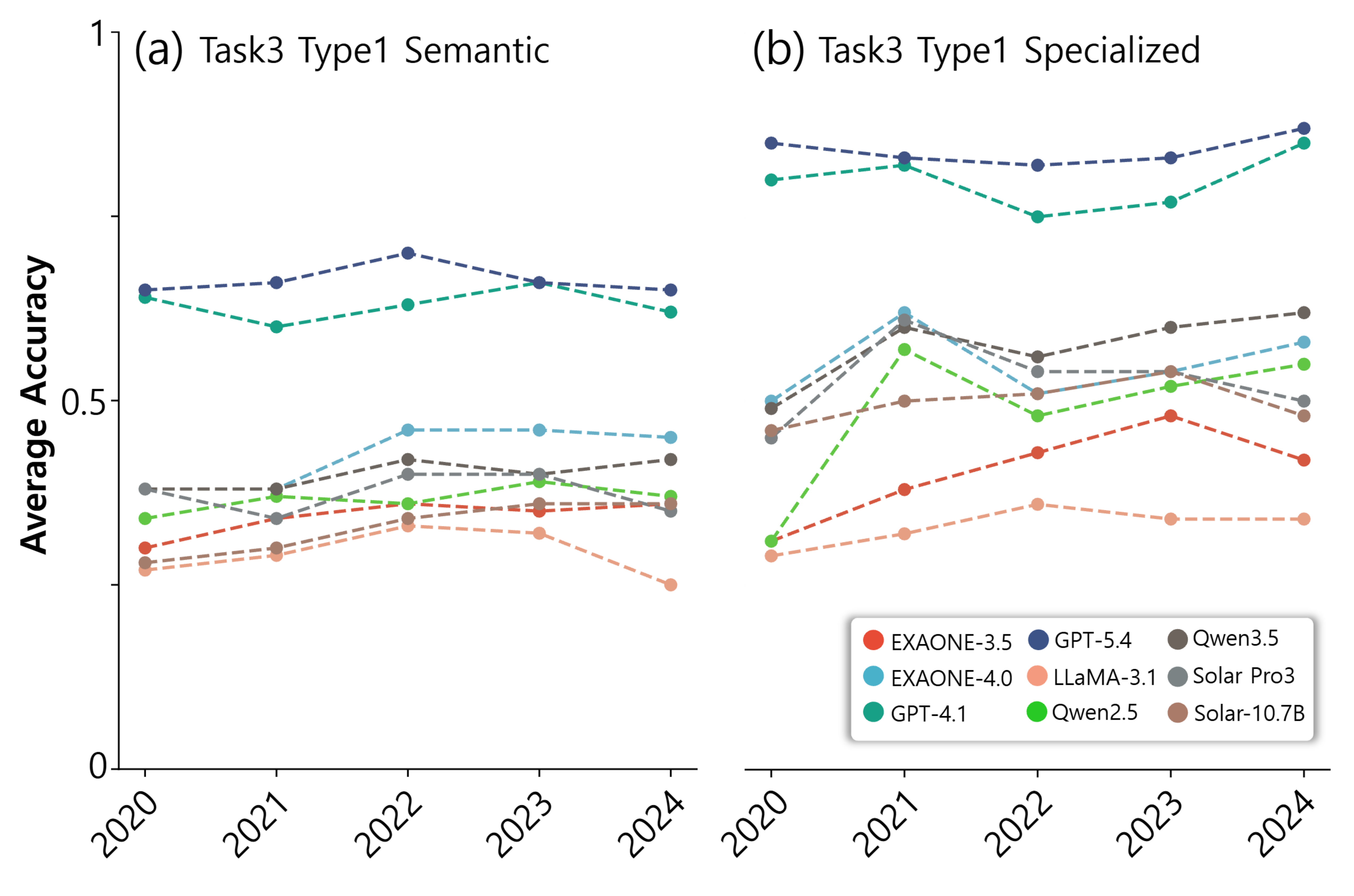}
    \caption{Year-wise performance distributions for Task 3 type 1. (a) Semantic category. (b) Specialized domain }
    \label{fig:year-f2}
\end{figure*}

\begin{figure*}[ht]
    \centering
    \includegraphics[width=\textwidth]{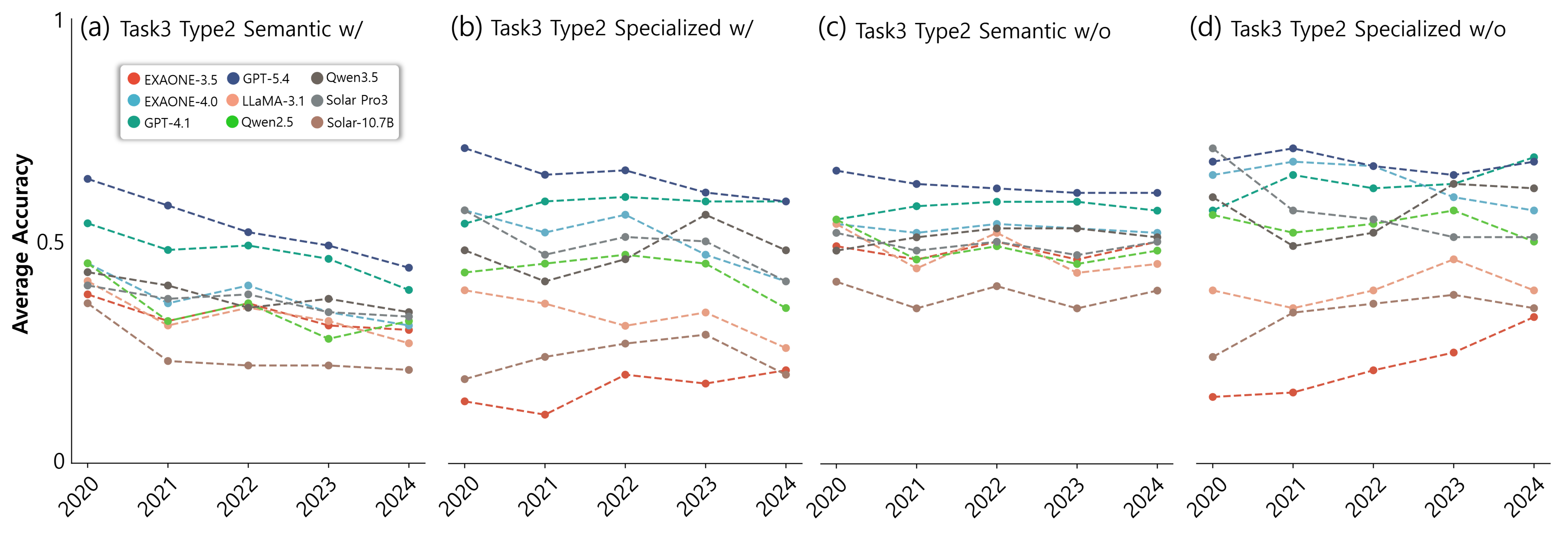}
    \caption{Year-wise performance distributions for Task 3 type 2. (a) Semantic category, w/ ex. (b) Specialized domain, w/ ex. (c) Semantic category, w/o ex. (d) Specialized domain, w/o ex.}
    \label{fig:year-f3}
\end{figure*}

In addition, year-wise performance distributions for Tasks 1 and 2, Task 3 Type 1, and Task 3 Type 2 are provided in Figures~\ref{fig:year-f1}, \ref{fig:year-f2}, and \ref{fig:year-f3}, respectively. Tasks 1 and 2 show an overall decline in performance for more recent neologisms, similar to Task 4. In contrast, Task 3 exhibits relatively little year-to-year variation in both the general-language and specialized-domain settings

\begin{figure*}[ht]
    \centering
    \includegraphics[width=\textwidth]{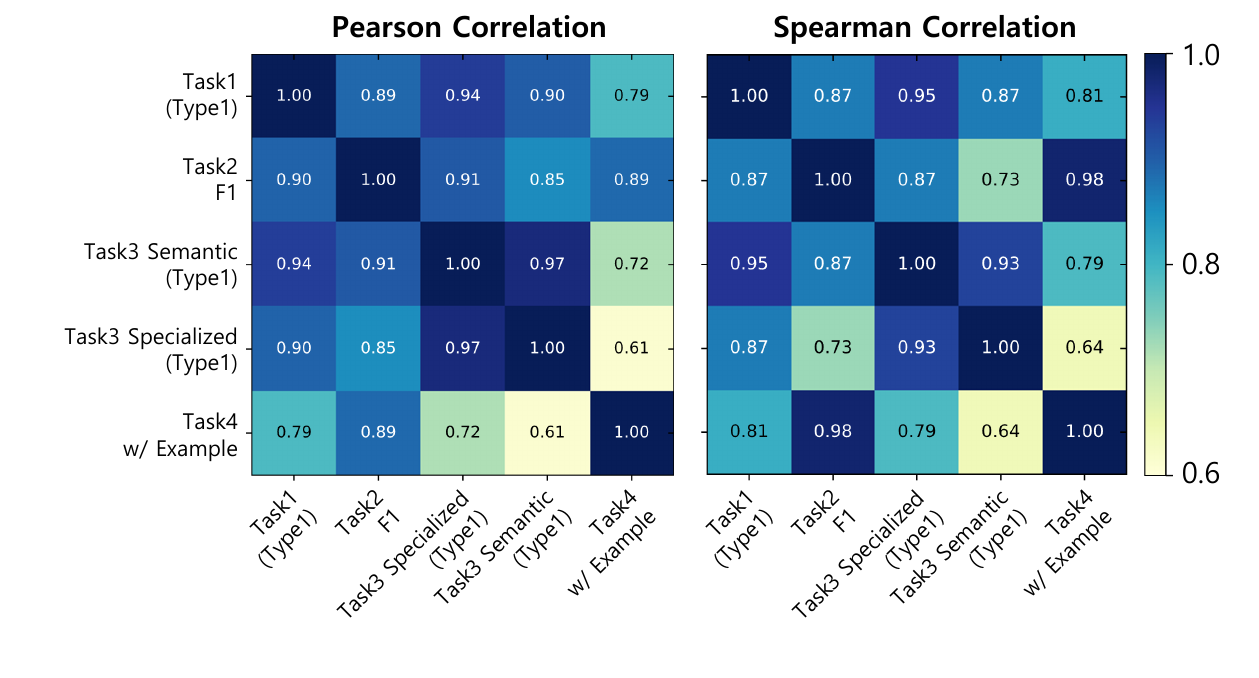}
    \caption{Correlation heatmap of task-wise model performance vectors}
    \label{fig:Task_correlation_heatmap}
\end{figure*}

Figure~\ref{fig:Task_correlation_heatmap} shows the Pearson and Spearman correlations of model performance across representative tasks. Most task pairs are positively correlated, and many exceed 0.7. However, the correlations are not uniformly high. In particular, Task 4 shows weaker correlations with some categorization tasks, suggesting that definition generation requires additional abilities (reconstructing the core meaning of a neologism and expressing it in dictionary style) that are not fully captured by multiple-choice or classification tasks.

These results support the multi-task design of KoNeoBench. High inter-task correlations indicate that the tasks measure a shared ability, namely Korean neologism understanding, while lower correlations and task-specific differences show that each task captures distinct linguistic abilities and error patterns. Rather than reducing neologism understanding to a single score, KoNeoBench evaluates it from complementary perspectives: contextual identification, source word identification, categorization, and definition generation.

\subsection{Performance Analysis on Task 4}
\label{app:C.analysis_task4}

\subsubsection{Gemini Scores vs. Human Scores}
\label{app:C.2.general}
To assess the reliability of Gemini-based evaluation, we also conducted human validation with a lexicography expert. The validation set consisted of definitions generated by GPT-5.4 in the setting with both the neologism and its usage example. We first sampled 45 items, selecting five from each Gemini total-score level from 2 to 10, and then added 205 more in proportion to the overall score distribution, for a total of 250 definitions. The expert evaluated them using the same rubric as in Table 9, and we compared the resulting scores with the Gemini scores at both the total-score and subcriterion levels.

\begin{figure*}[t]
    \centering
    \includegraphics[width=\textwidth]{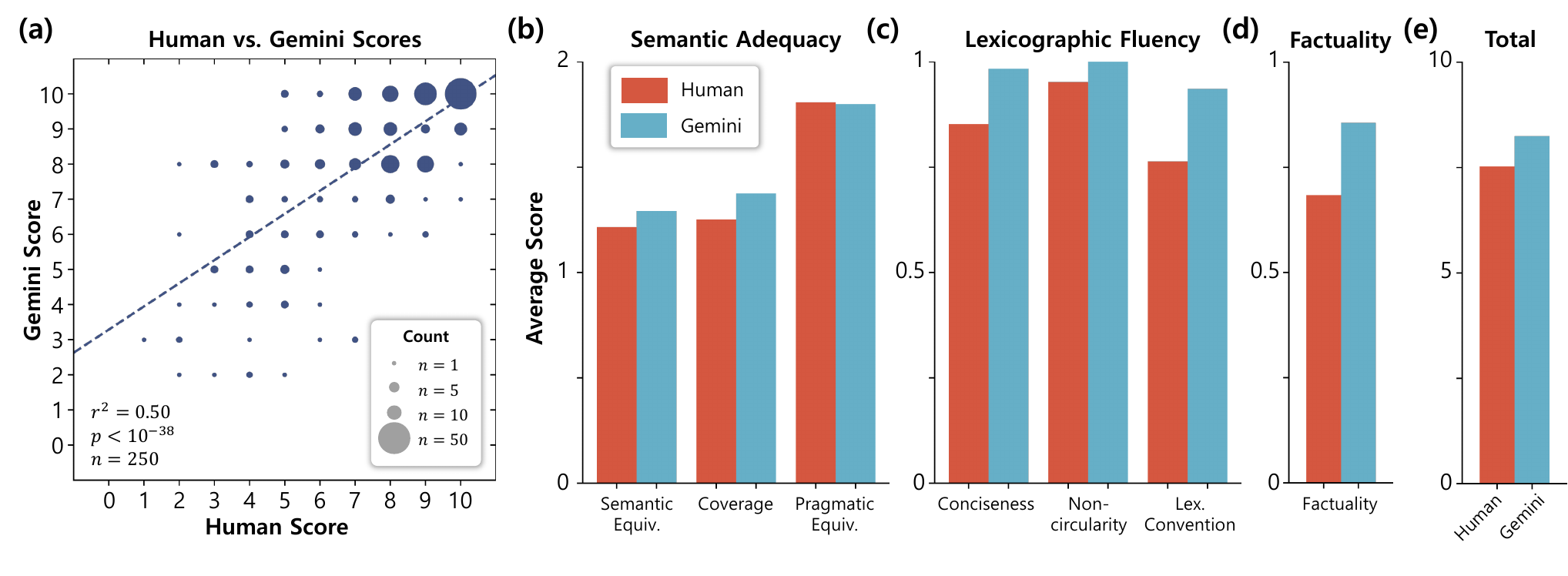}
    \caption{Task4. Comparison of human and gemini judge score}
    \label{fig:task4_human_vs_gemini}

\end{figure*}

Figure~\ref{fig:task4_human_vs_gemini} summarizes the comparison. Figure~\ref{fig:task4_human_vs_gemini}(a) shows a bubble plot of the total-score distributions, where each point represents a human–Gemini score pair and bubble size indicates the number of items. The two evaluations are positively correlated (r = 0.708), with linear regression results of $R^2 = 0.50, p < 10^{-38}$. This indicates that definitions scored highly by Gemini also tend to be scored highly by the expert. As shown in Figure~\ref{fig:task4_human_vs_gemini}(b), the discrepancy is relatively small for criteria related to semantic adequacy, suggesting that Gemini-based evaluation is reasonably consistent in assessing semantic quality. Gemini can therefore be used as a useful auxiliary measure of overall definition quality.

However, Gemini tends to assign slightly higher scores than the expert, as seen in the average subcriterion scores in Figure~\ref{fig:task4_human_vs_gemini}(b)-(e). This tendency is especially clear for conciseness, lexicographic convention, and factuality, where Gemini appears more lenient. This suggests that Gemini may judge formal quality, dictionary style, and factual errors more permissively than a human expert.

These results suggest that LLM-based evaluation is useful for comparing relative model differences in Task 4, but limited as an absolute measure of lexicographic quality. The Task 4 scores in the main text should therefore be interpreted as automatic scores for comparing models under the same conditions, rather than as absolute quality judgments. We also note that the human validation covers only 250 definitions generated by GPT-5.4, not all models or outputs. Even so, the analysis suggests that Gemini-based evaluation is reasonably consistent with expert judgment and can serve as a scalable approximate method for comparing large sets of generated definitions.

\subsubsection{Distribution of Scores by Subcriterion for Task 4}
\label{app:last}
\begin{figure*}[t]
    \centering
    \includegraphics[width=0.9\textwidth]
    {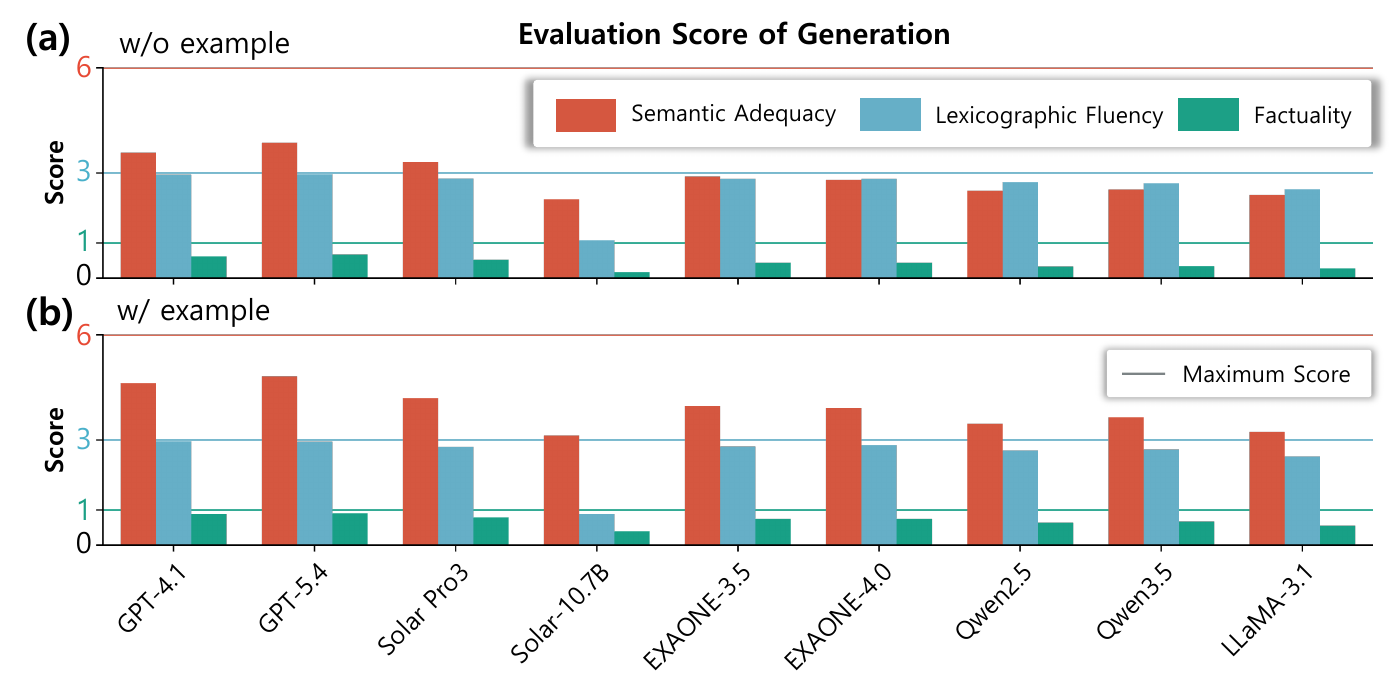}
    \caption{Average scores of generated definition.}
    \label{fig:generation scores}
\end{figure*}

Figure~\ref{fig:generation scores} shows the average subcriterion scores of each model on Task 4. We analyze definition generation from three perspectives: semantic adequacy, lexicographic fluency, and factuality. Figure~\ref{fig:generation scores}(a) shows the results when only the neologism is given, and Figure~\ref{fig:generation scores}(b) shows the results when a usage example is also provided.

Overall, most models improve on all three dimensions when usage examples are provided. The largest gain appears in semantic adequacy, suggesting that usage examples are crucial for inferring the actual meaning and context of Korean neologisms. Since many neologisms cannot be interpreted reliably from form alone, usage context helps models construct their core meaning more accurately. By contrast, differences in lexicographic fluency are relatively small, suggesting that LLMs can generate superficially dictionary-like definitions even without fully understanding the target meaning. This indicates that the main difficulty of Task 4 lies less in definitional form than in semantic accuracy. Scores on factuality also improve across all models, suggesting that usage examples help reduce hallucinated or unsupported content.

The subcriterion-level analysis of Task 4 clearly shows the importance of contextual information in Korean neologism definition generation. Even without usage examples, some models can produce relatively natural definitions, but performance is more stable with examples, especially in terms of semantic adequacy and factuality. In addition, the subcriterion-level analysis reveals differences that are not visible when definition generation is evaluated only by a single total score. These results suggest that Task 4 evaluates not just generation ability, but also neologism interpretation, context-based reasoning, and lexicographic description.

\section{Human Performance Evaluation}
\label{app:C.3.human_eval}

We conducted a human evaluation on four settings: Task 1 with particle-attached forms, Task 2, and the semantic and specialized-domain categories of Task 3 (Type 1). For each setting, 50 neologisms were sampled with consideration of their frequency distribution. Twenty native Korean speakers participated, including 10 linguistics majors and 10 non-majors. The LLMs were evaluated on the same subsets. The results are reported in Table~\ref{tab:human_evaluation_task}.

The comparison shows different patterns across tasks. On Tasks 1 and 2, linguistics majors performed better than both GPT-5.4 and non-majors. This may reflect the greater role of Korean morphological and word-formation knowledge in these tasks. The pattern was less consistent for Task 3. Linguistics majors performed better than non-majors on the semantic category setting, while GPT-5.4 performed better than both non-majors and the overall human group on the specialized-domain setting. Thus, the relative difficulty for humans and models differs depending on the type of knowledge tested.

We used paired Wilcoxon signed-rank tests over the 50 items to examine these differences. On Task 1, linguistics majors significantly outperformed GPT-5.4 ($W=175.0$, $p=.0217$) and non-majors ($W=82.5$, $p<.001$). The same pattern was observed on Task 2, where linguistics majors outperformed GPT-5.4 ($W=311.0$, $p=.0318$) and non-majors ($W=38.5$, $p<.001$). On the semantic category setting of Task 3, linguistics majors performed significantly better than non-majors ($W=132.5$, $p<.001$). On the specialized-domain setting, GPT-5.4 performed significantly better than non-majors ($W=270.5$, $p=.0374$) and the overall human group ($W=285.0$, $p=.0370$). No other comparisons were significant. Because the evaluation was conducted on a relatively small subset, a larger human evaluation remains for future work.

\begin{table*}[t]
\centering
\small
\setlength{\tabcolsep}{7pt}
\begin{tabular}{lcccc}
\toprule
\textbf{Model / Group}
& \textbf{Task 1}
& \textbf{Task 2}
& \textbf{Task 3 (Semantic)}
& \textbf{Task 3 (Special)} \\
\midrule
GPT-4.1
& 46.00\% & 47.05\% & 68.00\% & 72.00\% \\
GPT-5.4
& 60.00\% & 51.01\% & 70.00\% & 82.00\% \\
Solar Pro3-12B
& 32.00\% & 35.90\% & 44.00\% & 48.00\% \\
Solar-10.7B
& 28.00\% & 6.80\% & 36.00\% & 46.00\% \\
EXAONE-3.5-7.8B
& 48.00\% & 33.72\% & 38.00\% & 40.00\% \\
EXAONE 4.0-32B
& 48.00\% & 24.27\% & 50.00\% & 50.00\% \\
Qwen2.5-7B
& 34.00\% & 15.20\% & 30.00\% & 44.00\% \\
Qwen3.5-9B
& 46.00\% & 19.84\% & 44.00\% & 58.00\% \\
LLaMA 3.1-8B
& 30.00\% & 13.98\% & 32.00\% & 44.00\% \\
\midrule
Human average ($n=20$)
& 68.80\% & 56.80\% & 64.70\% & 75.40\% \\
Linguistics majors ($n=10$)
& 75.00\% & 63.40\% & 68.40\% & 76.20\% \\
Non-majors ($n=10$)
& 62.60\% & 50.19\% & 61.00\% & 74.60\% \\
\bottomrule
\end{tabular}

\caption{
Human and model performance on matched 50-item subsets.
}
\label{tab:human_evaluation_task}
\end{table*}

